%% file: sn-article.tex
\documentclass[pdflatex,sn-mathphys-num, oneside]{sn-jnl}%

\usepackage{graphicx}%
\usepackage{multirow}%
\usepackage{amsmath,amssymb,amsfonts}%
\usepackage{amsthm}%
\usepackage{mathrsfs}%
\usepackage[title]{appendix}%
\usepackage{xcolor}%
\usepackage{textcomp}%
\usepackage{manyfoot}%
\usepackage{booktabs}%
\usepackage{algorithm}%
\usepackage{algorithmicx}%
\usepackage{algpseudocode}%
\usepackage{listings}%
\usepackage{float}
\usepackage{booktabs}   
\usepackage{longtable}  
\usepackage{array}

\theoremstyle{thmstyleone}%
\theoremstyle{thmstyletwo}%

\theoremstyle{thmstylethree}%

\begin{document}


\title[]{A Multi-Agent Framework for Zero-Dimensional Reduced-Order Model Planning}


\author[1,2]{\fnm{Bingteng} \sur{Sun}}\email{sunbingteng@iet.cn}

\author[1,2]{\fnm{Hao} \sur{Yin}}\email{yinhao@iet.cn}
\equalcont{These authors contributed equally to this work.}

\author[1,2]{\fnm{Yiling} \sur{Chen}}\email{chenyiling@iet.cn}
\equalcont{These authors contributed equally to this work.}

\author[1,2]{\fnm{Renjie} \sur{Xiao}}\email{xiaorenjie@iet.cn}

\author[1,2]{\fnm{Lei} \sur{Xie}}\email{xielei@iet.cn}

\author[1,2]{\fnm{Shanyou} \sur{Wang}}\email{wangshanyou@iet.cn}

\author[1,2]{\fnm{Ruonan} \sur{Wang}}\email{wangruonan@iet.cn}

\author[1,2]{\fnm{Shubao} \sur{Chen}}\email{chenshubao@iet.cn}

\author[1,2]{\fnm{Qingzong} \sur{Xu}}\email{xuqingzong@iet.cn}

\author[6]{\fnm{Lin} \sur{Lu}}\email{llu@sdu.edu.cn}

\author*[1,2,3,4,5]{\fnm{Qiang} \sur{Du}}\email{duqiang@iet.cn}

\author[1,2]{\fnm{Junqiang} \sur{Zhu}}\email{zhujunqiang24@sina.com}

\affil[1]{\orgdiv{Advanced Gas Turbine Laboratory}, \orgname{Institute of Engineering Thermophysics, Chinese Academy of Sciences}, \orgaddress{\city{Beijing}, \postcode{100190}, \country{China}}}

\affil[2]{\orgdiv{National Key Laboratory of Science and Technology on Advanced Light-duty Gas-turbine}, \orgaddress{\city{Beijing}, \postcode{100190}, \country{China}}}

\affil[3]{\orgdiv{}, \orgname{University of Chinese Academy of Sciences}, \orgaddress{\city{Beijing}, \postcode{100190}, \country{China}}}

\affil[4]{\orgdiv{}, \orgname{Qingdao Institute of Aeronautical Technology}, \orgaddress{\city{Qingdao}, \country{China}}}

\affil[5]{\orgdiv{}, \orgname{Nanjing Future Energy System Research Institute}, \orgaddress{\city{Nanjing}, \state{Jiangsu}, \postcode{211135}, \country{China}}}

\affil[6]{\orgdiv{School of Computer Science and Technology}, \orgname{Shandong University}, \orgaddress{\street{72 Binhai Road}, \city{Qingdao}, \postcode{266237}, \state{Shandong Province}, \country{China}}}


\abstract{High-end complex equipment generally adopts multi-dimensional design workflows, among which the development of zero-dimensional reduced-order models (0D ROMs) serves as the core of the entire design process.
Currently, this model planning process heavily relies on manual experience, facing bottlenecks such as a limited exploration space for network topologies and long iteration periods.
Even with the introduction of traditional optimization methods like the Genetic Algorithm (GA), the process is often restricted to local parameter optimization.
In recent years, Large Language Model (LLM) agents, equipped with powerful decision-making and tool-use capabilities, have demonstrated the ability to rapidly explore massive sample spaces, bringing new opportunities to engineering design.
Although the Chain of Thought (CoT) and Reason and Act (ReAct) frameworks significantly improve the reliability of agent reasoning, and Retrieval-Augmented Generation (RAG) technology effectively overcomes domain knowledge barriers, a single agent still falls short when dealing with the long-horizon and highly coupled nature of complex 0D ROM planning tasks.
To address these challenges, this paper proposes a Zero-dimensional reduced-order model CO-Planning framework based on a multi-Agent architecture (Z-COPA).
This framework is equipped with a Symbolic Action Graph Engine (SAGE) and a MILP-Guided Navigation (MGN) optimizer.
Its core innovation lies in proposing a dedicated graph representation method to accurately describe the 0D flow network topology, thereby innovatively converting the traditional empirical model planning process into a rigorous graph structure optimization problem.
We validated the forward and inverse design capabilities as well as generalization performance of Z-COPA on three types of cases, including two real aircraft engine secondary-air system design cases, two IEEE power-distribution reconfiguration benchmarks, and two water-distribution network design benchmarks.
The results show that our multi-agent system achieves superior task completion quality, obtaining the best performance in both forward and reverse design of air systems. It also cuts the active power loss of the IEEE 69-bus distribution network by 55.46\% and reduces the cost for the Two-Loop water network by 90.45\%.
The proposal of Z-COPA disrupts the traditional 0D model planning paradigm, providing a brand-new technical approach for exploring a broader topological space and achieving highly automated, globally optimal air system architectures.}


\maketitle

\input{1_intro}
\input{2_Results}

\input{3_Discussion}
\input{4_Methods}

\section*{CRediT authorship contribution statement}
\textbf{Bingteng Sun:} Methodology, Software, Validation, Visualization, Writing-original draft. 
\textbf{Hao Yin:} Data curation, Software, Writing - review \& editing.
\textbf{Yiling Chen:} Data curation, Supervision, Writing - review \& editing.
\textbf{Renjie Xiao:} Conceptualization, Supervision,  Writing-review \& editing.
\textbf{Lei Xie:} Data curation, Writing - review \& editing.
\textbf{Shanyou Wang:} Writing - review \& editing.
\textbf{Ruonan Wang:} Writing - review \& editing.
\textbf{Shubao Chen:} Writing - review \& editing.
\textbf{Qingzong Xu:} Writing - review \& editing.
\textbf{Lin Lu:} Writing - review \& editing.
\textbf{Qiang Du:} Data curation, Supervision, Writing - review \& editing.
\textbf{Junqiang Zhu:} Conceptualization, Supervision, W Writing-review \& editing.

\section*{Declaration of competing interest}
The authors declare that they have no known competing financial interests or personal relationships that could have
appeared to influence the work reported in this paper.

\section*{Data availability}
Data will be made available on request.

\section*{Acknowledgements}
The authors wish to acknowledge the financial support of Excellence Research Group Program (ERGP, the 
former Basic Science Center Program) (Grant No.52488101) and the project supported by Young Scientists Fund (A Class, Grant No. 52525603), and the Shandong Provincial Natural Science Foundation (Grant No. ZR2024JQ011).


\bibliography{sn-bibliography}

\newpage
\appendix
\onecolumn
\input{5_Appendix}

\end{document}

%% file: 1_Intro.tex
\section{Introduction}
\label{sec:intro}

  






In modern engineering, the efficient design and optimization of complex systems usually involves 0-dimensional, 1-dimensional, 2-dimensional and 3-dimensional modeling and simulation processes, which require a delicate balance between computational cost and physical accuracy\cite{Culubret2024,Roknujjaman2026,Duronio2020}.
To address this, the engineering community widely adopts a highly abstract and representative modeling approach: simplifying complex three-dimensional (3D) physical systems into 0-dimensional (0D) low-order reduced models \cite{Rubio2026}.
By discretizing spatially continuous physical processes into networks composed of nodes and links, these 0D models focus on the macroscopic transfer of mass, momentum, and energy within the system, thereby significantly enhancing computational efficiency while preserving core physical characteristics \cite{Pinardi2025}.
Such 0D engineering design problems, abstracted from complex physical phenomena, are ubiquitous and highly representative across various disciplines.
For instance, the dynamic planning of urban water networks and pipeline systems in municipal and energy engineering \cite{Wang2023}, as well as the design of aero-engine Secondary Air Systems(SAS) and thermodynamic cycles in the aerospace sector \cite{Yan2026ANC}\cite{Zou2024}, fundamentally rely on such 0D network models. Consequently, the 0D optimization design problems abstracted from complex real-world challenges across diverse fields serve not only as the cornerstone for early-stage configuration selection, performance evaluation, and global optimization of engineering systems, but also stand as a universal and critically important core research topic in modern engineering design.

To address the aforementioned 0D optimization design problems, traditional solution strategies can be broadly classified into three categories. 
The first and currently most mainstream paradigm is manual empirical design, which relies heavily on domain-specific expert knowledge, heuristics, and the personal experience of engineers \cite{wu2017combining, Liu2020}. 
The second category comprises gradient-free optimization algorithms, such as Genetic Algorithms (GA) and Bayesian Optimization (BO) \cite{Sangroula2022, Gupta1999, Liang2025}, which explore the design space without requiring derivative information.
The third category involves gradient-based optimization algorithms \cite{https://doi.org/10.48550/arxiv.2603.09197, Cembrano2006, antaki2003piping}, which leverage mathematical gradients to efficiently converge upon optimal solutions in continuous design spaces.
However, despite their widespread application, a fundamental limitation persists in both gradient-free and gradient-based approaches: they function merely as passive computational tools.
These algorithms lack the cognitive capacity to actively comprehend the physical connotations, underlying mechanisms, and implicit design requirements of the engineering problems \cite{Clay2025, Chiarello2024, Daareyni2025}.
Consequently, they require extensive human intervention to formulate objective functions and constraints, which severely hinders the transition towards fully autonomous and intelligent engineering design.

Traditional optimization algorithms have inherent limitations in semantic understanding and intent recognition, making it difficult to directly translate high-level human design intentions into low-level engineering parameters.
In recent years, Large Language Models (LLMs), with their exceptional natural language processing and logical reasoning capabilities, have gradually emerged as a crucial bridge to close this gap \cite{zhao2023survey}.
Driven by this trend, the single-agent problem-solving paradigm was first applied to complex tasks.
Beyond the foundational Chain-of-Thought (CoT) \cite{https://doi.org/10.48550/arxiv.2201.11903} prompting, researchers have successively proposed classic frameworks such as ReAct (Reasoning and Acting) \cite{yao2023react}, Toolformer \cite{schick2024toolformer}, and Reflexion \cite{shinn2024reflexion}.
These advancements endow single agents with the abilities to invoke external tools, interact with environments, and perform self-reflection and error correction.
However, when confronted with highly complex engineering design and optimization tasks, single agents often fall short in global task planning and long-horizon reasoning.
They are prone to reasoning interruptions or error accumulation, making it challenging for them to independently tackle complex problems with deep domain dependencies \cite{https://doi.org/10.48550/arxiv.2305.02412}.

To break through the capability bottlenecks of single agents, the research paradigm has begun to evolve towards multi-agent collaborative architectures.
Recently, multi-agent frameworks represented by CAMEL \cite{li2023camel}, ChatDev \cite{Qian2024}, AutoGen \cite{wu2023autogen}, and MetaGPT \cite{hong2023metagpt} have emerged.
By employing refined role definitions, task decomposition, and Standard Operating Procedures (SOPs), these frameworks allow multiple specialized agents to engage in role-playing, debate, and cross-validation, thereby significantly improving the efficiency of planning and solving complex problems.
A representative case study, SciSciGPT, proposes a "Research Manager-Expert" collaborative architecture and an Agent Capability Maturity Model for the field of scientometrics, enabling the automated generation of full-process research prototypes—ranging from literature comprehension and data extraction to iterative visualization\cite{Shao2025}.
Although multi-agent architectures demonstrate immense potential in logical orchestration and task breakdown, a pure text-based agent network relying solely on LLMs remains insufficient for rigorous engineering optimization.
On the one hand, the parametric knowledge of LLMs carries the risk of hallucinations and struggles to cover deep, domain-specific expert experience.
On the other hand, they lack an intrinsic perception of physical laws, making it impossible to guarantee the physical feasibility of the generated solutions.

Therefore, to achieve truly reliable intelligent design, it is imperative to introduce Retrieval-Augmented Generation (RAG) technology to dynamically acquire and align with domain-specific expert knowledge.
From the early standard RAG \cite{lewis2020retrieval} to Self-RAG \cite{asai2023self}, which features on-demand retrieval and self-critique capabilities, and further to Graph Retrieval-Augmented Generation (GraphRAG) \cite{edge2024from}, which can handle complex entity relationships, the evolution of RAG technologies provides agents with increasingly precise external knowledge anchors.
Building upon this, deeply coupling physical simulation engines to provide agents with real physical feedback and closed-loop validation \cite{ma2023eureka} has become an inevitable trend.
SimulRAG enables the retrieval of numerical information from scientific simulators, translates numerical simulations into textual context via a generalized retrieval interface, and employs a statement-level verification method based on uncertainty estimation\cite{https://doi.org/10.48550/arxiv.2509.25459}.
This integrated architecture is emerging as the frontier paradigm for solving complex engineering optimization problems.
Meanwhile, ontology-driven RAG technology is evolving toward fully collaborative intelligence\cite{Venkatesh2026}.
It not only leverages LLMs for the semi-automatic construction, maintenance and link prediction of ontologies, but also explores adopting ontologies as an external reasoning space\cite{https://doi.org/10.48550/arxiv.2403.08345}.
With path search algorithms such as Think-on-Graph guiding models to perform complex multi-hop logical reasoning, this technology delivers verifiable and interpretable decision support for rigorous fields including biomedicine and engineering design\cite{Liu2025}.
Specifically, Hyper-RAG captures complex entity associations beyond-pairwise correlations—within a knowledge base through hypergraph modeling\cite{Feng2026}.

\begin{figure*}[htp]
    \centering
    \includegraphics[width=1.0\textwidth]{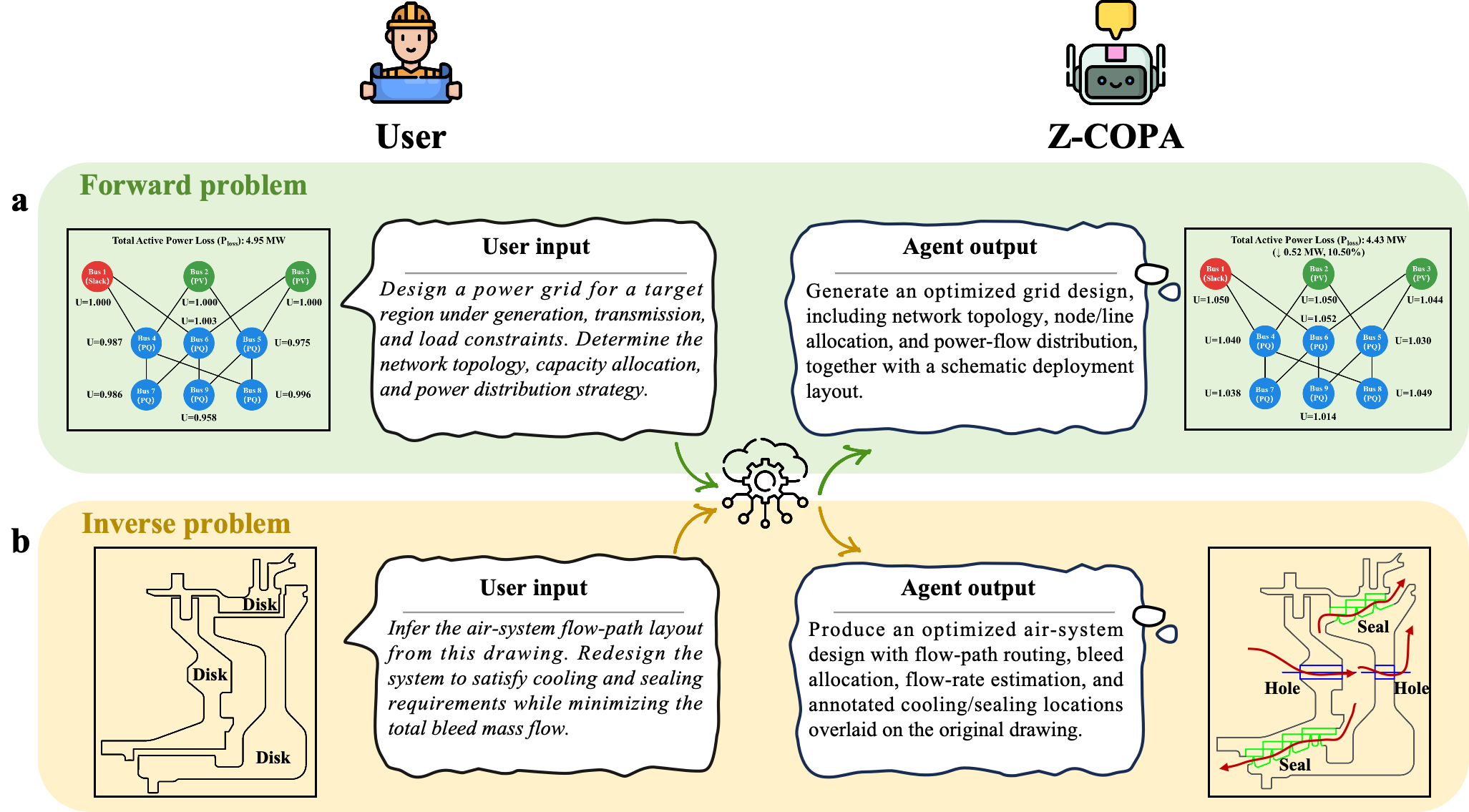}
    \caption{\textbf{The Z-COPA workflow illustrating the dual-mode operation:} \textbf{(a)} forward design path, where topology and parameters are given and performance is evaluated; \textbf{(b)} inverse design path, where performance targets are specified and the system searches for feasible topology--parameter configurations.}
    \label{fig:1}
\end{figure*}

To address the above problems, we introduce Z-COPA, an autonomous multi-agent framework designed to tackle both forward and inverse 0-dimensional reduced-order model (0D ROM) planning problems.
As illustrated in Fig.~\ref{fig:1}, we formulate 0D ROM planning as a unified constrained topology-parameter co-optimization problem. Let a candidate design be defined as $x = (\mathcal{T}, \mathcal{P})$, where $\mathcal{T}$ denotes the flow-network topology and $\mathcal{P}$ represents the associated continuous engineering parameters. Given the target performance $\mathcal{R}_{target}$ and physical constraints $g(x) \le 0$, Z-COPA aims to find the optimal configuration $x^*$ that minimizes the mismatch $\mathcal{L}$ between the simulator's physical response $\mathcal{S}(\mathcal{T}, \mathcal{P})$ and the desired target:

\[
x^* = \arg \min_{\mathcal{T}, \mathcal{P}} \mathcal{L}(\mathcal{S}(\mathcal{T}, \mathcal{P}), \mathcal{R}_{target}) \quad \text{s.t.} \quad g(\mathcal{T}, \mathcal{P}) \le 0
\]

Within this formulation, the forward problem represents a specific deterministic case where the topology $\mathcal{T}$ is completely predefined by input evidence (e.g., existing CAD drawings). The agents solely focus on iteratively refining the parameter vector $\mathcal{P}$ over the fixed graph to safely meet performance thresholds. Conversely, the inverse problem addresses blank-slate scenarios by activating the full multi-agent generative loop. It autonomously constructs valid topologies $\mathcal{T}$ from scratch and jointly optimizes $\mathcal{P}$, transforming traditional manual trial-and-error into an explicit, automated graph-structure exploration. Furthermore, $\mathcal{L}$ is solved using the 0D ROM graph physics-informed neural network solver (Z-GPSolver).
%

Our contributions are as follows:
\begin{itemize}[nosep]
    \item Pioneering Multi-Agent Framework: We propose Z-COPA, the first multi-agent collaborative architecture tailored specifically for 0D ROM planning. This framework operationalizes LLM agents to automate the complex network design and topology exploration process under strict physical and engineering constraints.
    \item Symbolic Action Graph Engine (SAGE): We introduce a novel ontology-driven multimodal Retrieval-Augmented Generation module: SAGE. This component empowers the agents to deeply comprehend complex physical problems, accurately identify key optimization variables, and intelligently guide the optimization direction through domain-specific knowledge grounding.
    \item MILP-Guided Navigation (MGN) Algorithm: We develop the MGN algorithm, a hybrid optimization strategy that bridges discrete topological search with continuous parameter tuning. By leveraging Mixed-Integer Linear Programming (MILP) to navigate the complex non-convex search space and Z-GPSolver for efficient solution cases, MGN effectively avoids local optima and accelerates convergence.
    \item Empirical State-of-the-Art: Extensive evaluations demonstrate that Z-COPA achieves state-of-the-art performance across diverse 0D ROM planning tasks. It significantly outperforms traditional manual design heuristics and strong numerical baselines in both optimization quality and success rate, while substantially reducing the wall-clock time required for complex system design.
\end{itemize}

%% file: 2_Results.tex
\section{Results and discussion}
\label{sec:2_results&discussion}

\subsection{Overview of Z-COPA}





\begin{figure*}[htbp!]
    \centering
    \includegraphics[width=1\textwidth]{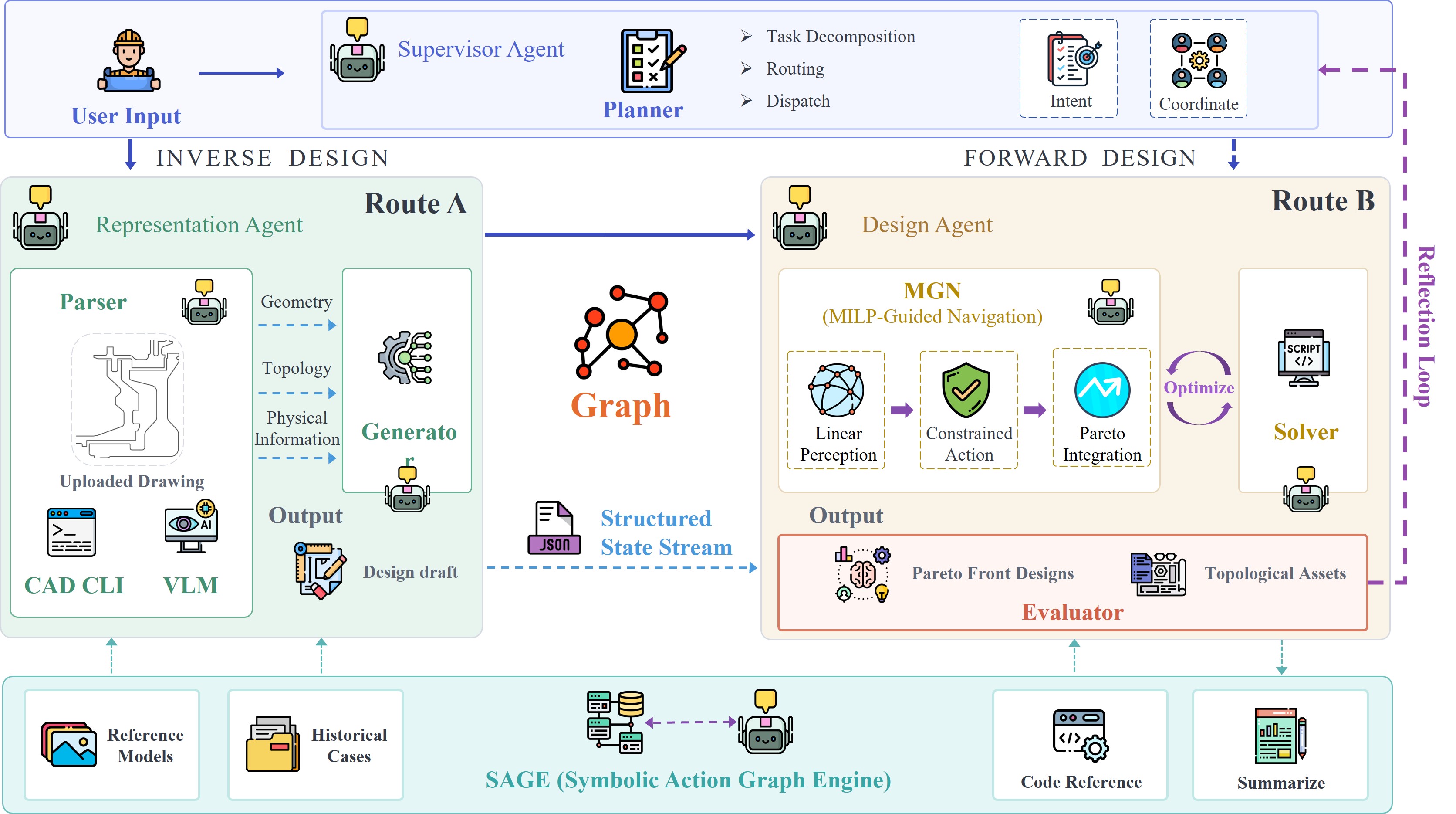}
    \caption{\textbf{Architecture of the proposed Z-COPA framework.}
    The three primary agents (Supervisor, Representation, Design) are coordinated by a meta-level harness. Solid lines indicate the forward design data flow; dashed lines indicate the inverse design loop. The SAGE module supplies domain knowledge to all agents.}
    \label{fig:2}
\end{figure*}

This work proposes Z-COPA, as shown in Fig.~\ref{fig:2}, a universal LLM-driven multi-agent intelligent framework dedicated to complex zero-dimensional optimization design scenarios, including engine air system collaborative design, regional power grid layout optimization, and urban integrated water pipeline network scheduling design. 
To fundamentally break through the above practical bottlenecks in industrial design, Z-COPA integrates three core original innovations throughout the whole process: modular collaborative multi-agent architecture, ontology-driven multi-modal RAG knowledge parsing mechanism, and MGN optimization algorithm, which jointly support the whole-link autonomous design from demand perception to scheme iteration and performance verification.

The overall functional deployment of Z-COPA relies on a unified scheduling multi-agent system, which independently matches two standardized engineering design paths according to different initial design conditions, and realizes full-process intelligent coordination of all functional modules and professional engineering tools:

\textbf{\textit{Forward planning.}}
This design path requires an existing initial zero-dimensional design scheme.
The system can directly activate the independent optimization branch of Route B and use the established initial design parameters as the input benchmark for the entire process.
Relying on the efficient iterative optimization capability of the MGN hybrid optimization algorithm, the embedded professional core engineering tool zero-dimensional graph PINN solver is invoked in real time throughout the iteration process.
The solver efficiently calculates real-time performance responses of multiple groups of perturbed design parameters, feeds back accurate performance evaluation indicators to the MGN algorithm, realizes rapid convergence and high-precision polishing of design parameter groups under fixed topological conditions, and finally outputs optimal design parameter configurations that meet all engineering performance indicators.

\textbf{\textit{Inverse planning.}}
This design pipeline targets blank engineering design scenarios without any initial zero-dimensional design prototypes.
Relying on the core capabilities of ontology-driven multimodal RAG, the agent collaboratively invokes a suite of professional engineering tools, including a full-condition query tool for DWG drawings and a visual feature perception model for engineering drawings.
It extracts three types of information from raw engineering drawings and historical design databases: geometric boundary conditions, topological constraint relationships, and multi-dimensional physical information containing implicit engineering requirements.
The agent then independently generates feasible initial design schemes with complete topological frameworks.
All structured perception data and prototype schemes are subsequently delivered to the downstream optimization Route B, enabling integrated autonomous operation covering full topological scheme generation, overall layout matching, and joint multi-objective parameter optimization.

All tool scheduling, data interaction and design logic iteration of Z-COPA are managed and constrained by the underlying engineering knowledge base engine.
The entire process is recorded in JSON format.
The whole framework forms a complete closed-loop automatic correction mechanism: if the performance detection link verifies that the output scheme fails to meet industrial standard thresholds and actual operating requirements, the system will automatically backtrack the design link, intelligently adjust RAG perception depth or MGN iteration intensity, and restart the design optimization cycle. Compared with traditional manual empirical design, Z-COPA effectively decouples the strong dependence on designer experience, realizes standardized, mechanism-aware and high-precision autonomous iteration of zero-dimensional engineering whole-process design, and has strong engineering application promotion value in the field of fluid machinery and energy power system design.

\textbf{Hardware and Software Configuration:}
All experiments were conducted on a local workstation equipped with an Intel Core i7-14700F processor, 32 GB of DDR5 system memory and an NVIDIA GeForce RTX 4060 Ti GPU with 8 GB of GDDR6 memory. The experimental system was implemented in Python 3.12 and used Microsoft AutoGen AgentChat 0.7.5 as the core multi-agent framework, with autogen-agentchat, autogen-core and autogen-ext all fixed at version 0.7.5. Owing to the limited local GPU memory, language and vision-language inference was performed through the remote Kimi API using the multimodal, tool-calling \texttt{kimi-k2.6} model. Tool-access permissions were isolated according to agent roles: optimization, retrieval and solution-verification tools were exposed only to the agents responsible for the corresponding functions, thereby reducing unnecessary tool exposure and constraining agent behaviour boundaries.

\textbf{Dataset:}
We evaluated the proposed method on six network-design cases spanning aero-engine secondary-air systems, power distribution systems and water distribution systems. The two SAS cases were derived from real CAD-driven secondary-air-system design workflows rather than synthetic graph instances. In the forward-design task, the CAD/VLM pipeline first extracts the airflow topology and converts it into a solver-ready request, after which optimization is performed mainly over design parameters on a fixed topology. In the inverse-design task, the same CAD/VLM-to-solver conversion flow produces a validated branch-region solver artifact, and the optimization jointly modifies topology and parameters under secondary-air-system physical objectives. These two cases represent realistic engineering airflow design settings in which geometry interpretation, component grounding, solver materialization and multi-objective feasibility constraints are tightly coupled.

For public benchmarks, we used two radial distribution-network reconfiguration cases, IEEE case33 and IEEE case69, and two water-distribution network design cases, Two-Loop and New York Tunnels. The IEEE case33 and case69 feeders are widely used distribution-network reconfiguration benchmarks from the Baran--Wu feeder family \cite{Baran1989}. Candidate power-grid solutions were represented as switching actions over normally closed sectional lines and normally open tie switches, and were evaluated under strict radiality, connectivity, AC power-flow convergence, voltage-bound and line-loading constraints, with active power loss as the primary objective. The Two-Loop benchmark is a classical least-cost water-distribution design problem \cite{Alperovits1977,Savic1997}, whereas the New York Tunnels benchmark is a larger water-network expansion problem commonly used to assess pipe-sizing and parallel-pipe design methods \cite{Savic1997}. Water-network candidates were evaluated through EPANET/WNTR hydraulic simulation, with discrete pipe-diameter or parallel-pipe decisions scored by construction cost, resilience where applicable, hydraulic convergence, source reachability and minimum pressure/head constraints. Together, these cases test whether the method transfers from real industrial CAD-derived SAS design tasks to established physical-network benchmarks with nontrivial topology, solver and constraint structure.

\textbf{Benchmark Tasks:}
We designed a two-tier evaluation mechanism comprising basic static tasks and functional dynamic tasks. 
Basic tasks examine the reliability of individual agents and tools, whereas functional tasks evaluate whether these capabilities can be coordinated into solver-backed co-planning workflows.

\textit{Basic tasks} test individual framework components:
\begin{enumerate}[nosep]
    \item \textit{Preprocessing.} The system is evaluated on its ability to convert heterogeneous engineering inputs into structured, model-ready representations. Depending on the task, these inputs may include drawings, diagrams, tabular specifications, existing network descriptions, or configured solver contexts. The evaluation covers component recognition, geometric and attribute extraction, spatial or connectivity-relation identification, boundary-condition parsing, and consistency checks between the perceived structure and the downstream zero-dimensional model.

    \item \textit{SAGE retrieval.} The retrieval module is evaluated on six categories of technical questions: terminology grounding, component-function explanation, parameter semantics, constraint interpretation, failure-mode diagnosis, and optimization guidance. Each category is assessed from two perspectives: agent-facing diagnostic retrieval and user-facing explanatory retrieval.

    \item \textit{Solver interface.} Solver-side agents are evaluated for numerical accuracy, convergence behavior, runtime efficiency, and API fidelity. These tests verify that solver requests, candidate artifacts, objective metrics, and validation reports can be exchanged without altering the physical meaning of the design task.
\end{enumerate}

\textit{Functional tasks} assess coordinated co-planning performance:
\begin{enumerate}[nosep]
    \item \textit{Forward design.} Given an engineering model, drawing, or configured solver-ready topology, Z-COPA performs preprocessing and context verification, constructs an auditable 0D ROM solver request, and executes a parameter-search--solver-validation loop. The goal is to improve task-specific objective metrics while preserving solver feasibility, boundary conditions, and physical constraints.

    \item \textit{Inverse design.} Given incomplete engineering inputs, target requirements, and boundary constraints, Z-COPA first uses ontology-guided retrieval and engineering perception tools to extract structured information about boundaries, topology, components, and requirements. It then exposes legal topology and parameter actions and performs topology--parameter co-optimization with solver-backed validation. If a candidate fails constraint or performance checks, the workflow backtracks to the perception or action stage.
\end{enumerate}

\textbf{Evaluation metrics: }To compare Z-COPA across heterogeneous engineering cases, we evaluate each
method along three dimensions:
\begin{itemize}[nosep]
    \item \textbf{Task completion:} Success Rate (SR) measures whether a run
    completes the required agent--tool workflow and returns a solver-validated
    design with objective values extracted from executable artifacts. We use
    this permissive criterion to compare methods at a common executable endpoint
    and to assess ReAct/CoT orchestration reliability rather than final design
    quality.

    \item \textbf{Solution quality:} Objective-wise Improvement measures the
    relative gain over the initial input design, using relative reduction for
    minimized objectives and relative increase for maximized objectives. The
    Total Score is computed as a weighted sum of the configured
    objective-improvement ratios.

    \item \textbf{Computational efficiency:} Runtime measures the elapsed
    wall-clock time required to complete the workflow and produce the final
    solver-backed result.
\end{itemize}
The Total Score is used only for reporting and cross-baseline comparison; it is
not used as the internal optimization objective for NSGA-II, MGN, or
domain-specific optimizers. Detailed metric definitions are provided in
Appendix~\ref{app:metrics}.

\textbf{Baselines:}
We compared Z-COPA with four ablated configurations to isolate the contributions of SAGE retrieval, reasoning format, multi-agent decomposition and optimization policy. These baselines assess whether the target tasks can be solved by a single monolithic agent, whether SAGE improves domain grounding and the identification of optimization variables, and whether changing the search policy affects the quality of solver-backed designs. In these zero-dimensional engineering tasks, conventional optimization algorithms provide the numerical search backbone, but they do not parse heterogeneous design inputs, identify domain-specific variables, coordinate solver validation or generate auditable design evidence on their own.
Therefore, we did not include these traditional algorithms in the comparison baselines, but instead focused on comparing two classic agent frameworks: ReAct and  Compact CoT.

\begin{enumerate}
    \item[(1)] \textbf{Single Agent + ReAct + NSGA-II:}
    A monolithic ReAct-style agent that receives the task query and executes the auditable optimization workflow using NSGA-II for candidate generation. The agent has no access to SAGE retrieval. This configuration serves as the single-agent, retrieval-free evolutionary optimization baseline.
    
    \item[(2)] \textbf{Single Agent + ReAct + SAGE + NSGA-II:}
    The same monolithic ReAct agent as in (1), augmented with SAGE retrieval for domain-knowledge grounding. The optimizer, solver interface, candidate-evaluation budget and final scoring protocol are identical to those in (1). This configuration isolates the effect of adding retrieval under the same single-agent ReAct setting.
    
    \item[(3)] \textbf{Single Agent + Compact CoT + SAGE + NSGA-II:}
    A monolithic agent with the same tool access as in (2), but using a compact structured-rationale prompting format rather than ReAct. The agent provides concise planning and evidence summaries before tool decisions, while retaining the same SAGE access, NSGA-II candidate-generation policy, solver interface and evaluation protocol. This configuration isolates the effect of the reasoning format under a retrieval-enabled single-agent setting.
    
    \item[(4)] \textbf{Multi-Agent + SAGE + NSGA-II:}
    A role-decomposed multi-agent configuration that separates planning, context analysis, candidate-batch reflection, design optimization, solver validation and final evidence auditing. SAGE retrieval is available as an optional domain-knowledge source. Candidate generation uses the same NSGA-II policy, solver interface and candidate-evaluation budget as the single-agent NSGA-II baselines. This configuration evaluates the effect of multi-agent decomposition while keeping the optimizer family fixed.
    
    \item[(5)] \textbf{Multi-Agent + SAGE + MGN:}
    This configuration uses the multi-agent workflow in (4), but replaces NSGA-II with MGN for candidate generation, using the MGN configuration reported in Appendix~\ref{app:mgn_configuration}. MGN can search mixed topology--parameter candidates through GA-style population search with MILP-guided recombination or local branching. All other components are held fixed, isolating the effect of the optimizer family.
\end{enumerate}

\subsection{Case Studies}





To further validate the utility of Z-COPA, we applied it to three real-world zero-dimensional engineering design settings, including: 
(1) aero-engine SAS direct forward optimization planning, 
(2) aero-engine SAS inverse topology-parameter planning, 
and (3) physical-network design optimization covering power-grid reconfiguration and water-distribution design.

\subsubsection{SAS Direct forward optimization design}
    
The Forward Design Case evaluates whether the proposed multi-agent workflow can convert an uploaded secondary air system drawing into a solver-backed multi-objective design optimization process. Starting from the SAS design drawing, the system extracts components, flow paths and converts them into structured data, and then initiates a forward-design task to search for improved system configurations. The optimization seeks a Pareto trade-off among seal leakage, total bleed-flow consumption and pressure nonuniformity, while preserving solver convergence, source--sink connectivity, seal pressure margin and critical boundary-flow constraints. This case therefore tests both the physical quality of the optimized design and the ability of the agent workflow to plan, retrieve domain evidence, construct a solver-ready network and execute a constrained optimization loop.

\begin{figure*}[htp!]
\centering
\includegraphics[width=1\textwidth]{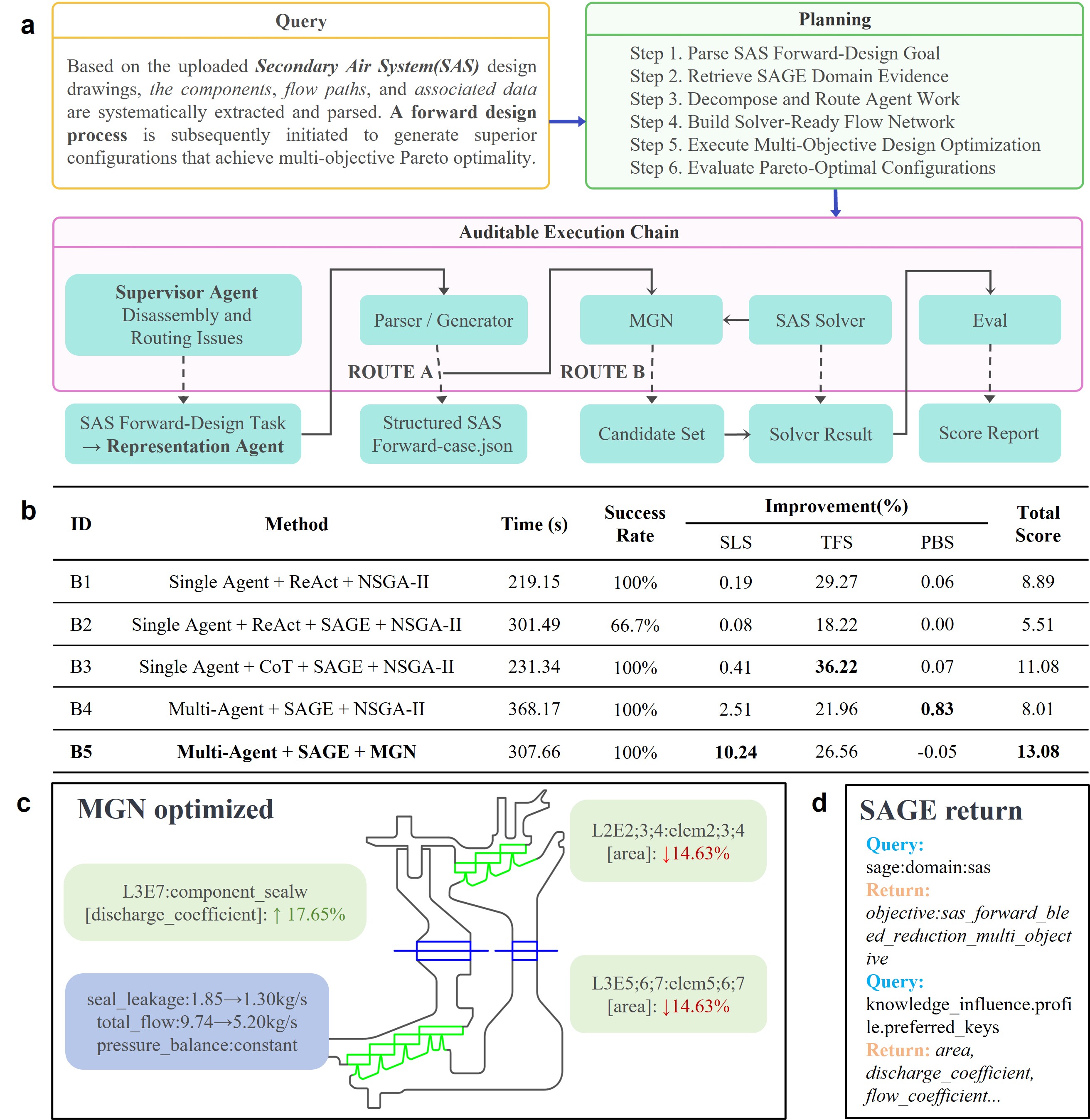}
\caption{
\textbf{Workflow and results of the Forward Design Case.}
\textbf{(a)} Input query and Planner-generated workflow. The query asks the system to parse the uploaded SAS drawings, extract components and flow paths, and initiate a forward-design process for multi-objective Pareto optimization. The planning trace contains six steps: parsing the SAS forward-design goal, retrieving SAGE domain evidence, decomposing and routing agent work, building a solver-ready flow network, executing multi-objective design optimization, and evaluating Pareto-optimal configurations.
\textbf{(b)} Quantitative results of the compared baselines, including runtime, success rate, objective-specific improvement scores and the reporting-time total score.
\textbf{(c)} MGN-optimized design evidence, showing the modified SAS configuration, parameter-level changes and the corresponding solver-backed objective values.
\textbf{(d)} SAGE retrieval evidence used by the agent workflow. The retrieval queries are scoped to the SAS domain and return the objective family \texttt{sas\_forward\_bleed\_reduction\_multi\_objective}, as well as preferred optimization variables such as area, discharge coefficient and flow coefficient.
}
\label{fig:forward_design_case}
\end{figure*}

As shown in Fig.~\ref{fig:forward_design_case}a and detailed in Appendix~\ref{app:agent_workflows}, the workflow begins with a natural-language query that specifies the uploaded SAS drawings as the design source and requests a forward multi-objective optimization process. The Planner first identifies the task as an SAS forward-design problem and decomposes it into six executable stages: goal parsing, SAGE evidence retrieval, agent-work routing, solver-ready network construction, multi-objective optimization and Pareto configuration evaluation. SAGE is used as a source of domain evidence rather than as a substitute for solver evaluation. In the retrieval trace shown in Fig.~\ref{fig:forward_design_case}d, the query is scoped to \texttt{domain:sas}; the returned objective family is \texttt{sas\_forward\_bleed\_reduction\_multi\_objective}, and the retrieved preferred keys indicate physically meaningful variables such as flow area, discharge coefficient and flow coefficient. These variables define the physically meaningful design space exposed to the solver-backed optimization process.

The quantitative comparison in Fig.~\ref{fig:forward_design_case}b reports objective-specific improvement scores for the three SAS metrics: \texttt{seal\_leakage\_kg\_s}, \texttt{total\_flow\_kg\_s} and \texttt{pressure\_cv}. The internal NSGA-II and MGN optimizers operate directly on this three-objective minimization vector without objective weights; the scalar total score is computed only for reporting and constraint-first final selection. The MGN design evidence in Fig.~\ref{fig:forward_design_case}c further shows how the selected configuration differs from the initial design at the parameter level and how these changes translate into solver-backed objective values. A run is counted as successful only when candidate generation, candidate selection and final solver validation are all supported by successful tool evidence, ensuring that the reported improvement is based on executable SAS design artifacts rather than text-only reasoning.

\subsubsection{SAS Inverse topology-parameter design}
    
The SAS inverse design case evaluates whether the proposed multi-agent workflow can start from a user query and uploaded SAS design drawings with specified inlet--outlet boundaries, and generate improved configurations through a constrained reverse-design process. Unlike the forward-design case, this task allows both continuous parameter updates and topology-editing actions within the target branch region. The optimization searches for Pareto-improving configurations over seal leakage, total flow consumption and pressure nonuniformity, while preserving solver convergence, source--sink connectivity, critical seal pressure margin and boundary-flow constraints. This case therefore tests whether the agent workflow can move beyond parameter tuning and produce solver-backed topology--parameter design alternatives.

\begin{figure*}[htp!]
\centering
\includegraphics[width=1\textwidth]{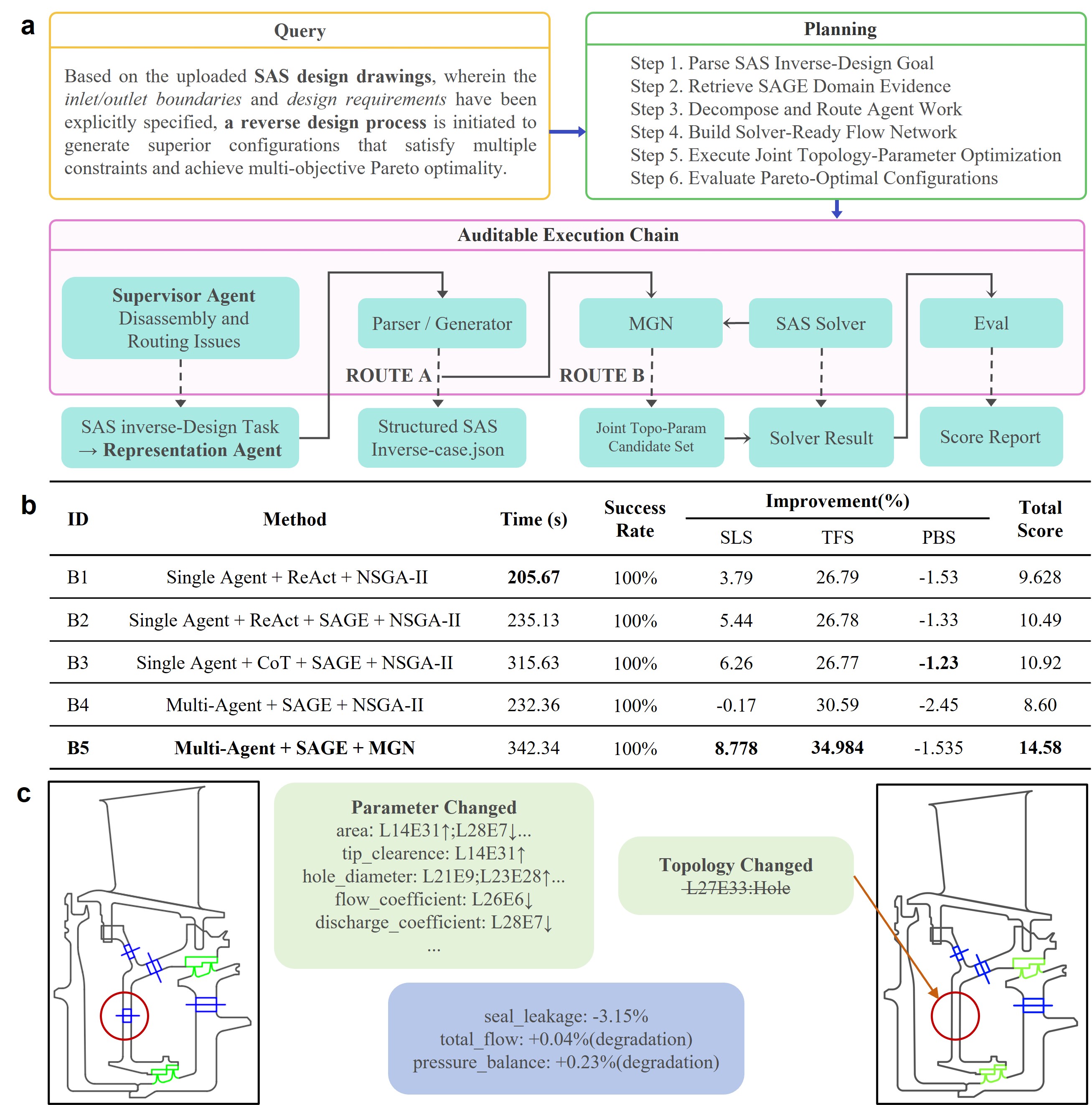}
\caption{
\textbf{Workflow and results of the SAS inverse topology--parameter design case.}
\textbf{(a)} User query and Planner-generated workflow. The workflow starts from a user query specifying the uploaded SAS drawings, inlet--outlet boundaries and reverse-design requirements. The Planner decomposes the task into six stages: parsing the SAS reverse-design goal, retrieving SAGE domain evidence, decomposing and routing agent work, building a solver-ready flow network, executing joint topology--parameter optimization, and evaluating Pareto-optimal configurations.
\textbf{(b)} Averaged quantitative results of the compared baselines, including runtime, success rate, objective-specific improvement scores and the reporting-time total score.
\textbf{(c)} MGN-optimized topology--parameter design evidence. The selected solver-backed design removes a redundant hole element in the branch region and jointly retunes neighbouring flow-area, clearance, discharge-coefficient and flow-coefficient parameters.
}
\label{fig:inverse_design_case}
\end{figure*}

As shown in Fig.~\ref{fig:inverse_design_case}a, the workflow starts from a user query rather than from a predefined optimization script. The query provides the uploaded SAS drawings, inlet--outlet boundary conditions and reverse-design requirements. The Planner identifies the task as an SAS inverse topology--parameter design problem and decomposes it into six executable stages: reverse-goal parsing, SAGE evidence retrieval, agent-work routing, solver-ready network construction, joint topology--parameter optimization and Pareto configuration evaluation. SAGE provides domain evidence for feasible variables, action directions and safe bounds, whereas final design validity is determined only by solver-backed candidate evaluation.

The quantitative comparison in Fig.~\ref{fig:inverse_design_case}b reports the same averaged objective-specific improvement scores as in the forward-design case: \texttt{seal\_leakage\_kg\_s}, \texttt{total\_flow\_kg\_s} and \texttt{pressure\_cv}. B5 achieves the highest overall score in this inverse-design setting because the MGN-based multi-agent workflow identifies a mixed topology--parameter solution rather than only retuning continuous parameters. In the selected design, the agent removes a redundant hole element in the branch region, substantially reducing seal leakage. This improvement entails a small trade-off: total flow and pressure nonuniformity increase slightly, but the final candidate remains solver-converged and satisfies all hard physical constraints. This behaviour is consistent with a constrained Pareto trade-off rather than a single-objective improvement.

The topology--parameter evidence in Fig.~\ref{fig:inverse_design_case}c shows the design mechanism underlying this result. The selected candidate removes the redundant hole element without adding new elements or edges, yielding a genuine topology simplification. Around this topology edit, the workflow retunes neighbouring branch parameters, including flow area, clearance, discharge coefficient and flow coefficient, to preserve feasibility. The resulting configuration reduces seal leakage by approximately 3.15\%, while the total-flow and pressure-nonuniformity objectives show only minor degradation. Thus, the B5 result demonstrates that the workflow can identify a solver-backed inverse-design alternative that simplifies the topology and improves the dominant leakage objective under multi-objective constraints.

A run is counted as successful only when candidate generation, explicit candidate selection and final solver validation are all supported by tool evidence. This requirement ensures that the reported inverse-design result corresponds to an executable SAS topology--parameter configuration rather than a text-only design suggestion.

\subsubsection{Physical Network Optimization Case}

The Power--Water Case tests whether the proposed agent-orchestrated workflow transfers from secondary air systems to two infrastructure domains with distinct physics: power-distribution reconfiguration and water-network pipe sizing. The evaluation covers two power-grid benchmarks, IEEE case33 and IEEE case69, and two water-network benchmarks, Two-Loop and New York Tunnels. In the power-grid tasks, the system searched for radial switching configurations that reduced active power loss while preserving voltage and loading feasibility. In the water-network tasks, it searched over pipe-sizing or parallel-pipe design decisions to reduce cost, or to improve cost--resilience trade-offs, while preserving hydraulic feasibility and pressure/head constraints. These cases tested whether SAGE-grounded domain knowledge and MGN-based solver-backed search could expose domain-specific design variables without requiring the language agent to implement the power-flow or hydraulic solver.

\begin{figure*}[htp!]
\centering
\includegraphics[width=1\textwidth]{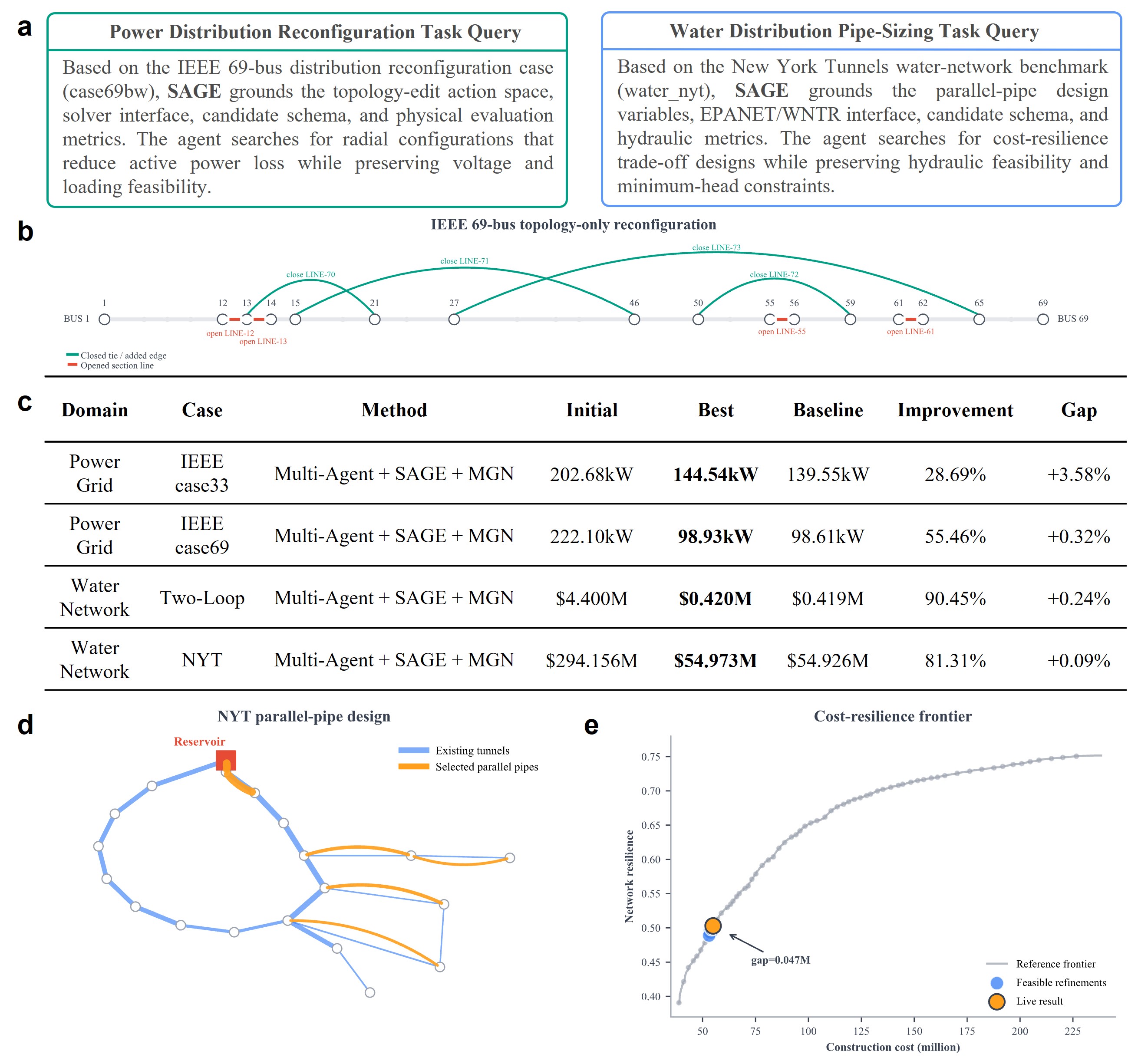}
\caption{
\textbf{Cross-domain optimization results for power-grid reconfiguration and water-network pipe sizing.}
\textbf{(a)} Task queries for the two highlighted domains. In the IEEE 69-bus power-distribution case, SAGE grounds the topology-edit action space, solver interface, candidate schema and physical evaluation metrics. In the New York Tunnels water-network case, SAGE grounds the parallel-pipe design variables, EPANET/WNTR interface, candidate schema and hydraulic evaluation metrics.
\textbf{(b)} Solver-backed power-grid reconfiguration evidence for the selected IEEE 69-bus design. The optimized topology closes tie lines and opens section lines to form a radial configuration with lower active power loss while maintaining voltage and loading feasibility.
\textbf{(c)} Quantitative comparison across four cross-domain benchmarks: IEEE case33, IEEE case69, Two-Loop and New York Tunnels. The workflow approaches the corresponding reference baselines in all four settings, with gaps of 3.58\%, 0.32\%, 0.24\% and 0.09\%, respectively.
\textbf{(d)} Water-network design evidence for the New York Tunnels case, showing the selected parallel-pipe additions over the existing tunnel network.
\textbf{(e)} Cost--resilience frontier for the New York Tunnels case. The selected live result lies close to the reference frontier, with a cost gap of 0.047M to the reference cost while satisfying hydraulic feasibility.
}
\label{fig:case3}
\end{figure*}

Fig.~\ref{fig:case3}a shows that the highlighted power and water tasks began from natural-language engineering queries, but the agent was not asked to solve the physical optimization problem directly. Instead, SAGE provided domain-grounded evidence about variables, constraints, solver interfaces and evaluation metrics, while MGN supplied a solver-backed search process. This division was needed because the two domains used different executable representations: radial switch actions with AC power-flow validation for IEEE distribution networks, and pipe-sizing or parallel-pipe design vectors with EPANET/WNTR hydraulic validation for water networks. The agent performed task interpretation, evidence retrieval, tool routing and candidate selection, whereas MGN performed candidate generation, materialization and fresh solver scoring.

Fig.~\ref{fig:case3}c summarizes the four public infrastructure benchmarks used in this transfer study. On the power-grid side, the workflow reached 144.54 kW on IEEE case33 and 98.93 kW on IEEE case69. These correspond to gaps of 3.58\% and 0.32\% to their reference baselines. On the water-network side, it reached 420,000 on Two-Loop and 54.973M on New York Tunnels, with gaps of 0.24\% and 0.09\%, respectively. These results show that the same SAGE--MGN protocol exposed and optimized distinct design variables across electrical and hydraulic network problems.

Fig.~\ref{fig:case3}b details the selected IEEE case69 reconfiguration. The search explored radial topology edits, closing candidate tie lines and opening section lines to preserve radiality. The resulting configuration reduced active power loss from 222.10 kW to 98.93 kW, a 55.46\% reduction, while retaining solver-backed checks for voltage and loading feasibility.

The New York Tunnels result is shown in Fig.~\ref{fig:case3}d,e. The selected live design added a small set of parallel pipes to the existing tunnel network and reached a construction cost of 54.973M with network resilience of 0.503. Relative to the initial 294.156M design, this corresponds to an 81.31\% cost reduction. The result remained within 0.09\% of the 54.926M reference cost. The frontier plot further shows that the live result lay close to the reference cost--resilience trade-off curve, with a cost gap of 0.047M to the reference cost. Hydraulic feasibility was evaluated through EPANET/WNTR. Together, these cases support the central claim that Z-COPA transfers an agent-orchestrated, solver-grounded optimization workflow across heterogeneous engineering domains, with executable optimization authority residing in solver-backed MGN search rather than text-only agent reasoning.

%% file: 3_Discussion.tex
\section{Conclusions}
\label{sec:3_conclusion}
This work presents Z-COPA, a novel multi-agent collaborative framework tailored specifically for 0D ROM planning of complex thermal engineering systems. Built upon LLM-driven multi-agent orchestration, ontology-driven multimodal RAG module and MGN optimizer, Z-COPA innovatively transforms traditional empirical 0D network design into rigorous graph structure optimization, and supports both forward parameter optimization and inverse topology-parameter co-planning tasks.

Extensive validations on three representative engineering scenarios, including real aero-engine SAS design cases, IEEE 69-bus power distribution network and Two-Loop/New York Tunnels water-distribution benchmarks, demonstrate that Z-COPA outperforms single-agent paradigms and conventional numerical optimization baselines remarkably in task completion rate, design solution quality and computational efficiency. 
Experimental results show that Z-COPA achieves the highest target total score of 13.08 in aero-engine air system forward design, and reduces seal leakage by 3.15\% in inverse topology–parameter optimization; meanwhile, it lowers the active power loss of IEEE 69-bus distribution network by 55.46\%, and reduces 90.45\% cost in the water-network Two-Loop task, demonstrating varying degrees of improvement compared to the optimal baseline.
It effectively expands the topological exploration space of 0D flow networks, realizes high automation and global optimal design of system architectures, and the embedded Z-GPSolver also significantly accelerates numerical solving speed while maintaining high physical simulation accuracy. Meanwhile, the framework decouples the strong reliance on designers’ empirical experience, breaking the bottlenecks of long iteration cycle and limited topology exploration existing in traditional manual and single optimization methods.

Despite the superior performance of Z-COPA in autonomous 0D ROM planning and cross-domain generalization, it still has a potential limitation: the inherent accuracy constraint of zero-dimensional simulation itself. Although Z-COPA can achieve reliable topology parsing, knowledge grounding, constrained optimization and closed-loop verification via multi-agent collaboration, ontology RAG and MGN algorithm, its final design quality fundamentally depends on the accuracy of underlying 0D modeling formulas. When key physical models such as the flow resistance and heat transfer of labyrinth teeth and rotating cavities in aero-engine air systems are overly simplified, the optimized topology and parameter configurations generated by Z-COPA may still deviate from the actual physical performance targets.

In future work, we will further iterate and optimize the Z-COPA framework. The core direction is to introduce 3D simulation information into the multi-agent architecture, construct a real-time correction mechanism for 0D reduced-order models by fusing high-fidelity 3D physical field data. This can compensate the accuracy deficiency of simplified 0D empirical models, further enhance the physical rationality and practical engineering applicability of optimized design results, and ultimately build a more complete, high-precision and closed-loop intelligent optimization system for complex thermal equipment network design.

%% file: 4_Methods.tex
\section{Methods}
\label{sec:4_method}







\subsection{Z-COPA Multi-Agent Architecture}

As shown in Fig.~\ref{fig:2}, we construct a multi-agent framework for engineering drawing comprehension and flow-network optimization design. The core idea is to decompose complex design tasks into three complementary processes: representation, execution, and reflection. The representation process recovers a computable graph-structured state from engineering drawings and design evidence; the execution process then performs physical solving, candidate generation, and optimization iteration on that graph structure; the reflection process continuously evaluates the quality of outputs from both representation and execution, deciding whether the task should proceed, roll back, or switch paths. Rather than using a large language model as a black box that directly generates the final design, we limit the role of agents to structured-state construction, judgment, tool invocation, and iterative correction, thereby improving the interpretability and traceability of the design process.

The system consists of a main agent and several sub-agents, whose implementation modules are organized as independently inspectable runtime, routing, solver, retrieval, benchmark and result components, as shown in Appendix~\ref{app:project_structure}. Following the manager protocol specified in Appendix~\ref{app:agent_prompts}, the main agent assumes cognitive control during design: it decides whether the current task should enter the representation path, the execution path, or trigger reflection, based on user objectives, drawing evidence, physical constraints, and current solver feedback. Sub-agents assume specific functions, including drawing parsing, network generation, candidate structure generation, and physical solving. Route~A on the left corresponds to the representation process, responsible for converting unstructured or weakly structured drawings into a computable graph-structured state; Route~B on the right corresponds to the execution process, responsible for candidate generation, physical solving, optimization search, and feedback iteration on the graph-structured state. The two paths are connected through a unified structured state stream, enabling the system to share the same intermediate representation between forward-design and reverse-design tasks. The reflection mechanism spans both paths, evaluating the acceptability of the design state after each execution cycle and redirecting the task back to Route~A for representational completion when necessary.

\subsubsection{Main agent and planner}

The main agent serves as the cognitive control centre of the framework, responsible for maintaining the global design state and coordinating task scheduling among sub-agents. It does not directly perform drawing parsing or numerical solving; instead, it delegates specific operations to the appropriate sub-agents based on the current design evidence.

The planner is the core decision module of the main agent. Its inputs comprise user design objectives, a snapshot of the current structured state stream, and evaluation reports from the reflection mechanism; its output is a task execution plan specifying which sub-agent to activate next, the intermediate objective to achieve, and fallback strategies upon failure. The planner maintains a dynamic task queue, prioritizing tasks according to representational adequacy, execution feasibility, and constraint satisfaction rates.

The planner operates at three levels. At the path-selection level, it determines whether the current task should enter Route~A for representational completion, Route~B for solving and optimization, or trigger the reflection mechanism for state assessment. At the resource-scheduling level, it coordinates concurrent execution of multiple sub-agents while avoiding write conflicts on the shared state stream. At the exception-handling level, when a sub-agent reports failure or anomalous output, the planner analyses the failure type and decides whether to retry, degrade, or switch strategy.

Separating the Planner from the main agent makes the cognitive control logic of the framework modular. The main agent maintains the global state and user objectives, whereas the Planner generates execution plans from the state stream and reflection outputs. Sub-agents then carry out specific parsing, solving and validation tasks, and write their results back to the shared state stream. This design decouples planning decisions, tool execution and state updates. When the design task or constraint conditions change, the system can adjust the Planner’s control strategy without rewriting the sub-agent implementations, thereby improving extensibility across different engineering design scenarios.

\subsubsection{Representation agents}

Route~A is the representation path in our framework. Its core objective is to recover a computable design state from engineering drawings, carried out collaboratively by the drawing parser and the network generator.

\textbf{Drawing parser.} The drawing parser is responsible for extracting geometric elements, spatial positions, inlets/outlets, cavity regions, connection relationships, and candidate components from CAD drawings, engineering annotations, and visual cues. Its inputs are raw CAD files (DWG/DXF format), engineering annotation layers, and optional scanned images; its output is structured JSON containing geometric primitives, hierarchy information, annotation text, spatial coordinates, and preliminary candidate regions for inlets/outlets/cavities. For CAD primitives with clear rules, the system preferentially uses deterministic parsing tools to read line segments, contours, holes, cavities, hierarchies, and annotation information. For objects with ambiguous semantics, visual language models are used for auxiliary recognition, such as flow boundaries, cavity divisions, local sealing structures, and component categories. The parser also outputs a source confidence score for each recognized object, providing reliability judgment for downstream agents.

The introduction of a visual language model enables the system to handle non-standard annotations, handwritten comments, and ambiguous boundaries. Furthermore, to provide the precise geometric and physical information required by the subsequent Action agent, a CAD secondary development command line is also provided as a tool for direct drawing query.

\textbf{Network generator.} The network generator constructs a unified graph-structure representation based on parsing results. Its inputs are structured JSON from the drawing parser, user-supplied design constraints, and similar-case topology references from SAGE; its output is an attributed design graph containing node sets (inlets, outlets, cavities, connection points, intermediate computation nodes), edge sets (pipes, holes, seals, labyrinths, and other flow branches), and attached geometric parameters, physical parameters, source confidence scores, and engineering constraint attributes on nodes and edges. It also outputs solver input drafts and reusable design artifacts.

The network generator transforms geometric and semantic information from parsing results into an intermediate state directly usable by solvers and optimization algorithms. Nodes and edges carry multi-dimensional attributes simultaneously, making design intent explicit and computable from its implicit expression in drawings. The generator supports incremental graph construction: when the drawing parser supplements new recognition results, the network generator can perform local updates on the existing graph structure rather than rebuilding the complete graph object.

We define this intermediate representation as an attributed design graph. Compared with directly having a large model generate complete solver inputs, the graph representation offers better interpretability, inspectability, and iterability. Sub-agents do not carry complete design objects in the dialogue context; instead, they read local structured state files and append fields, correct local relationships, or call backend composition tools as needed. This reduces context-length pressure and mitigates inconsistency problems caused by rewriting large objects.

The output of Route~A is not an ordinary text description but a continuously updatable structured design state, containing geometric and topological information, object sources, recognition confidence scores, parameter gaps, and computability judgments. The subsequent execution path performs solving and optimization directly on this structured state without reinterpreting the original drawing. Therefore, Route~A assumes the representational modeling function of the design object in the entire framework: it makes implicit engineering information in drawings explicit, structured, and transformed into a graph state consumable by computation modules.

\subsubsection{Action agent}

Route~B is the execution path in our framework. Its objective is to complete physical solving, candidate screening, and optimization iteration on the graph structure generated by Route~A, carried out collaboratively by the network optimizer and the network solver. They form a generat, solve, feedback, regenerate, optimization loop, tightly coupling design-space exploration with physical verification.

\textbf{Network optimizer.} The network optimizer is the candidate structure generation and screening sub-agent, whose core algorithm is MGN. It is responsible for generating diverse candidate topologies in the design space and quickly eliminating infeasible schemes through constraint screening. Its inputs are the graph structure generated by Route~A, boundary conditions, initial parameters, design objectives, e.g., pressure-drop limits, flow-distribution requirements, and manufacturing constraints e.g., minimum hole diameter, connectivity requirements. Its outputs are a filtered collection of candidate graph structures, each accompanied by constraint satisfaction rate, topological feature vector, and expected performance metrics, together with invalid schemes filtered by MILP and their violated constraint types for feedback correction.

Inspired by simulation-aware evolutionary strategies, MGN replaces blind random mutations with a mathematically grounded, three-phase collaborative optimization process: Global Linear Approximation, MILP-Guided Local Optimization, and Pareto-Driven Evolution.

\textbf{Phase 1: Global Linear Approximation (Perception).} 
Directly optimizing the black-box physical objectives over the entire domain is computationally prohibitive. Therefore, we partition the global design space into multiple small local sub-regions. Within each tightly defined sub-region, we assume the objective functions and physical responses vary linearly. We perform Cartesian sampling across these sub-regions to construct a local numerical derivative for each parameter. By evaluating these sample points, we obtain the first-order partial derivatives, enabling a first-order linear expansion around the local center \(\mathbf{x}_k\):
\[
    f(\mathbf{x}) \approx f(\mathbf{x}_k) + \nabla f(\mathbf{x}_k)^\top (\mathbf{x} - \mathbf{x}_k) = f(\mathbf{x}_k) + \sum_{i=1}^{N} \frac{\partial f}{\partial x_i}\Bigg|_{\mathbf{x}_k} (x_i - x_{k,i})
\]
where \(\mathbf{x}\) represents the continuous design variables, and \(\nabla f(\mathbf{x}_k)\) is the local gradient estimated via the Cartesian samples. 
By applying this linear approximation to all the divided sub-regions, we can process the optimization objective into an input acceptable to subsequent MILP, and because the partial derivatives of the optimization objective with respect to each parameter are calculated based on a large number of sampling points, our calculations are reliable, and subsequent optimizations make full use of this gradient information.

\textbf{Phase 2: MILP-Guided Local Optimization (Action).} 
Leveraging the linear expressions derived in Phase 1, we formulate a MILP problem for each sub-region. This module rapidly screens and optimizes the candidates by simultaneously tuning continuous parameters \(\mathbf{x}\) and discrete topological indicators \(\mathbf{z}\) (where \(z_j \in \{0, 1\}\) denotes the existence or connectivity of a specific structural edge/node). The MILP formulation effectively eliminates physically infeasible structures, e.g., those violating connectivity, boundary conditions, or manufacturing constraints by embedding them as linear constraints:
\[
    \min_{\mathbf{x}, \mathbf{z}} \quad \nabla f(\mathbf{x}_k)^\top \mathbf{x} + \mathbf{c}^\top \mathbf{z}
\]
\[
    \text{subject to} \quad \mathbf{A}\mathbf{x} + \mathbf{B}\mathbf{z} \leq \mathbf{b}, \quad \mathbf{z} \in \{0, 1\}^M
\]
where \(\mathbf{A}\) and \(\mathbf{B}\) are constraint matrices defining the physical and topological boundaries, and \(\mathbf{b}\) represents the constraint limits. Solving this MILP yields the optimal deterministic sample point \((\mathbf{x}^*, \mathbf{z}^*)\) for each local interval, ensuring that computational resources are not wasted on invalid topologies.

\textbf{Phase 3: Pareto-Driven Evolution (Integration).} 
Instead of relying on standard random initialization or naive genetic operators, the high-quality, mathematically verified optima generated from the MILP module are injected as the elite initial population into the NSGA-II evolution module. NSGA-II then performs non-dominated sorting and crowding distance calculations to search for the global Pareto front. This step systematically balances conflicting multi-objective metrics—such as minimizing pressure drop, maximizing flow uniformity, and reducing structural complexity—ultimately achieving significantly faster convergence and higher solution quality than traditional random crossover and mutation strategies.

MILP screening intervenes at the generation stage, eliminating invalid candidates before solving and significantly improving computational efficiency. The coupling of NSGA-II with graph structures enables optimization directly in the topological space rather than searching in a fixed parameter space, thereby supporting variable-topology design. Furthermore, constraint-violation information from the network optimizer can flow back directly to Route~A, triggering the network generator to repair the graph structure and forming a cross-path feedback loop.
Relevant case settings are provided in Appendix~\ref{app:mgn_configuration}.

\textbf{Network solver.} 
The network solver functions as the physical computation sub-agent, responsible for performing numerical computations on each candidate graph structure to obtain verifiable physical responses. The primary innovation of this sub-agent lies in its native coupling with Z-GPSolver, which operates as a closed-loop learned solver assistant. In contrast to traditional approaches that necessitate manual preprocessing, the proposed framework directly compiles the candidate graph structures, alongside their associated geometric and physical parameters as well as boundary conditions, into standardized numerical inputs.

\begin{figure*}[htp!]
    \centering
    \includegraphics[width=1\textwidth]{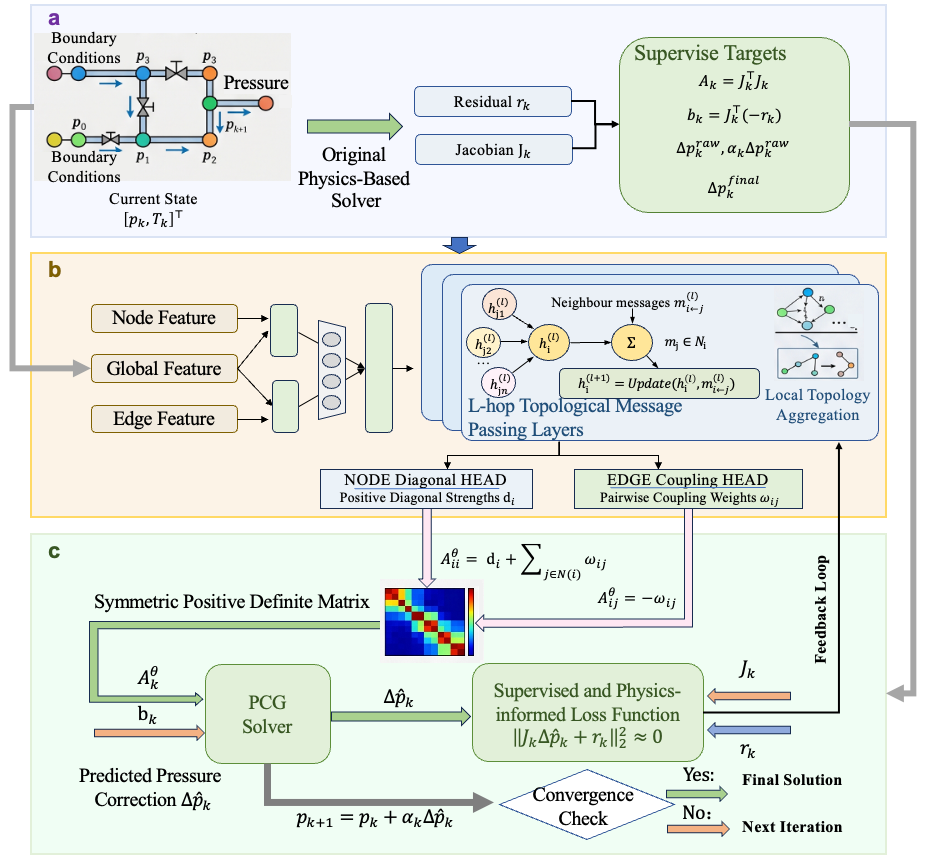}
    \caption{\textbf{The overall framework of Z-GPSolver.} \textbf{(a)} Data preparation within the physical-numerical solver. The original physics-based solver computes the residual $r_k$ and Jacobian $J_k$ from current states and boundary conditions to generate supervised targets. \textbf{(b)} Learned graph matrix builder. Graph neural networks with L-hop topological message passing aggregate node, edge, and global features to predict diagonal strengths and pairwise coupling weights. \textbf{(c)} Linear solve and Newton update. These predictions construct a SPD matrix for a preconditioned conjugate gradient (PCG) solver to compute the pressure correction $\Delta \hat{p}_k$. A physics-informed loss function guides the subsequent Newton update and convergence check, driving the feedback loop.}
    \label{fig:6}
\end{figure*}

As illustrated in Fig.~\ref{fig:6}, Z-GPSolver is a general graph-based framework for nonlinear physical-system solvers. At the $k$-th nonlinear iteration, given the current state $[p_k, T_k]^\top$ (where $p_k$ and $T_k$ denote the coupled physical states, such as pressure and temperature) and the boundary conditions, the underlying physics-based solver evaluates the nonlinear residual $r_k$ and, when required, the local linearization information such as the Jacobian matrix $J_k$. The correction step is then written in the generic symmetric positive definite (SPD) form
$$
A_k \Delta p_k = b_k,
$$
where $A_k$ and $b_k$ may be constructed directly from the physical model or obtained by augmenting a physics-based system with learned graph-structured corrections. In a common normal-equation setting, one may use
$$
A_k = J_k^\top J_k, \qquad b_k = J_k^\top (-r_k),
$$
although Z-GPSolver is not restricted to this specific instantiation. These primary variables are further utilized to construct the supervised targets shown in Fig.~\ref{fig:6}a, including the raw correction $\Delta p_k^{\mathrm{raw}}$, the relaxation-scaled correction $\alpha_k \Delta p_k^{\mathrm{raw}}$, and the final targeted response $\Delta p_k^{\mathrm{final}}$.

The graph neural network encodes node, edge, and global features and performs multi-layer message passing over the system topology. The hidden representation of node $i$ at layer $l+1$ is updated as
$$
h_i^{(l+1)}=\mathrm{Update}\!\left(h_i^{(l)}, \sum_{j\in \mathcal{N}(i)} m_{i\leftarrow j}^{(l)}\right),
$$
where $m_{i\leftarrow j}^{(l)}$ denotes the message passed from neighboring node $j$, and $\mathcal{N}(i)$ denotes the local topological neighborhood of node $i$. Based on the resulting embeddings, the network predicts the quantities needed to construct a structure-aware SPD surrogate. Specifically, it outputs the positive diagonal strengths $d_i$ and the pairwise coupling weights $\omega_{ij}$ to assemble the learned SPD matrix $A_k^\theta$, whose elements are explicitly defined as:
$$
A_{ii}^\theta = d_i + \sum_{j \in \mathcal{N}(i)} \omega_{ij}, \qquad A_{ij}^\theta = -\omega_{ij}.
$$
The resulting learned SPD system is then solved by PCG to obtain the model-assisted correction $\Delta \hat{p}_k$, which is used in a relaxed Newton-type update
$$
p_{k+1}=p_k+\alpha_k \Delta \hat{p}_k.
$$
Rather than serving as an unconstrained black-box predictor, the learning module is trained together with physics-consistency objectives by minimizing the residual-based loss function:
$$
\mathcal{L}_{\mathrm{res}}=\|J_k \Delta \hat{p}_k + r_k\|_2^2.
$$
This constraint explicitly enforces the physics-informed condition $\|J_k \Delta \hat{p}_k + r_k\|_2^2 \approx 0$ within the feedback loop, ensuring that the predicted correction remains consistent with the original nonlinear governing equations. After each update, the solver performs the standard convergence check and either accepts the new state or proceeds to the next nonlinear iteration. In this way, Z-GPSolver functions as a structure-aware acceleration module for repeated linear correction solves, reducing computational cost while preserving the overall physics-based solution process and maintaining high solution fidelity. Relevant case settings and validation details are provided in Appendix B.2 and C.

\subsubsection{Evaluation and reflection mechanism}

The evaluation process is embedded in the reflection iteration of Route~B rather than modeled as an independent agent chain. After each solving completion, the execution agent compares and reflects on current candidates by combining topological connectivity, physical responses, constraint satisfaction rates, objective function values, and historical optimization records. If a scheme satisfies design objectives, Pareto-front designs, topology assets, and structured reports are output. If there is excessive leakage, abnormal pressure drop, convergence failure, or topological constraint conflicts, the problem is written back to the structured state stream, triggering the next round of parameter adjustment, edge generation, topological repair, or candidate screening. When solver feedback reveals graph-structure deficiencies or insufficient boundary conditions, the main agent can redirect the task back to Route~A for representation completion. Reflection is therefore not a one-off judgment at the end of the process, but a feedback control mechanism that runs through the entire optimization process.

\subsubsection{Structured state stream}

Route~A and Route~B are connected through the Structured State Stream, whose solver-ready JSON-like graph representation is shown in Appendix~\ref{app:json_structure}. This state stream preserves not only the graph structure itself but also boundary conditions, parameter sources, tool invocation results, solver results, optimization history, and reflection records. Through this state stream, agents do not need to repeatedly copy complete design objects in natural-language context but perform incremental read--write on existing structured states. This reduces information-loss risks from long contexts and ensures that every design modification and computation result has a clear source.

\subsubsection{SAGE knowledge engine}
SAGE provides the memory and knowledge layer of the system. Its construction, including the ontology-guided organization of documents, solver code and case graphs, is described in Section~4.2; here we focus on its role within the agent architecture. SAGE stores reference models, historical cases, code references and task summaries, and makes this engineering knowledge retrievable by the agents during both design routes. In Route~A, it supports drawing comprehension and parameter completion: when the parser encounters semantically ambiguous regions, agents can retrieve recognition records from similar historical drawings to help determine cavity boundaries and component categories. In Route~B, it supports solver repair and optimization strategy selection: when the network solver fails to converge, agents can retrieve successful cases with similar topologies to obtain evidence for parameter adjustment and boundary-condition configuration. SAGE does not replace the computational role of the sub-agents or the solver; instead, it provides external evidence and traceable engineering experience for local agent decisions.

\subsubsection{Agent Harness runtime support}

The Agent Harness serves as the runtime support layer, maintaining task states, message passing, tool permissions, artifact versions, and execution logs between the main agent and sub-agents. The main agent uses the Harness to determine whether the current task is in the representation or execution stage and delegates specific operations to corresponding sub-agents. When sub-agents execute tasks, they do not maintain isolated contexts individually but read and write around the shared Structured State Stream. The Harness also provides lifecycle management for sub-agents, including launch, monitoring, termination, and fault recovery, ensuring that an anomaly in a single agent does not compromise overall system stability. This design helps avoid state-drift problems common in multi-agent coordination and ensures that every design modification, tool invocation, and solver result can be traced.



\subsection{Ontology-driven multimodal RAG} 

\begin{figure*}[htbp!]
    \centering
    \includegraphics[width=1\textwidth]{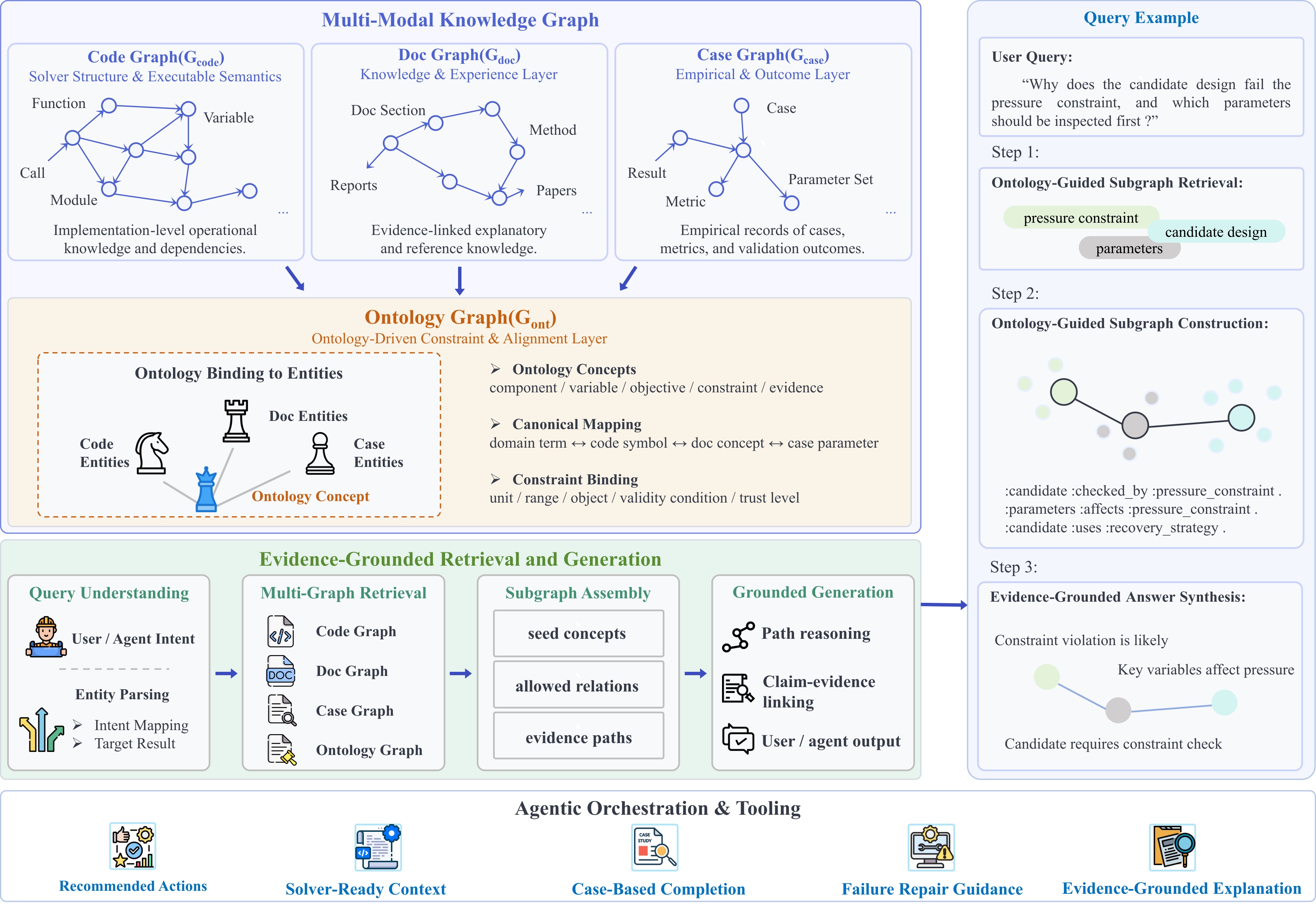}
    \caption{\textbf{SAGE architecture for ontology-driven engineering retrieval and reasoning.}
    SAGE organizes engineering knowledge into a unified multimodal knowledge graph. The ontology graph defines domain concepts, variables, constraints and relation schemas, and serves as the semantic layer that binds the other graphs. The code graph exposes solver implementations, formulas, parameters and dependencies; the document graph stores evidence extracted from technical materials; and the case graph records empirical designs, parameter settings and outcomes. During ingestion, entities from documents, code and cases are linked to ontology concepts to create shared semantic anchors across heterogeneous sources. At query time, candidate entities are parsed from the user or agent request, retrieved across graphs and expanded into an ontology-constrained subgraph. The resulting evidence is assembled through subgraph paths and synthesized into either human-readable explanations or structured agent context for solver inspection, case-based completion, failure repair and evidence-grounded decision support.}
    \label{fig:rag}
\end{figure*}

We developed SAGE as an ontology-driven retrieval-augmented generation system for organizing engineering knowledge and supporting agent-facing reasoning. As illustrated in Fig.~\ref{fig:rag}, SAGE supports both human users and autonomous engineering agents by converting heterogeneous engineering resources into semantically aligned, queryable graph structures.

SAGE integrates several types of engineering knowledge, including text documents, PDF manuals, figures or image-derived content, structured case files and solver source code. These sources are organized as four connected views within a unified multimodal knowledge graph. Within this unified graph, the ontology graph defines domain concepts, aliases, variables, constraints and relation schemas. The document graph stores entities and evidence extracted from technical documents. The code graph represents the white-box structure of the solver, including implementations, formulas, parameters and program dependencies. The case graph stores historical or generated design cases, including topology, objectives, parameter settings and convergence metadata. Together, these graph views provide complementary representations of the same engineering problem.

\subsubsection{Ontology-guided graph binding}

The core mechanism of SAGE is ontology-guided graph binding, which aligns entities from documents, code and cases through shared ontology concepts. During ingestion, documents, cases and code repositories are parsed into chunks, entities, relations and metadata. Extracted terms, code symbols, parameter names and case attributes are matched against ontology concepts and aliases. The resulting ontology bindings provide shared semantic anchors across otherwise heterogeneous sources. For example, a physical quantity may be linked simultaneously to its textual explanation in documents, its implementation variable or formula path in solver code, and its observed behavior in historical cases. This binding allows SAGE to retrieve and compose evidence at the concept level, rather than relying only on surface string matching.

SAGE stores these representations in a layered backend. A primary PostgreSQL store records documents, entities, relations, ontology definitions and bindings. Qdrant stores dense vector embeddings for semantic retrieval. Neo4j stores typed graph nodes and edges for graph traversal and subgraph construction. Redis is used for caching repeated query results. Embedding, reranking and entity extraction are accessed through a model gateway. This separation makes the retrieval pipeline auditable and robust: each channel can be inspected independently, and keyword or graph retrieval remains available when vector indexing or generation services are unavailable.

\subsubsection{Query processing and subgraph construction}

At query time, SAGE first identifies candidate entities, ontology labels and domain constraints from the user or agent query, thereby converting a natural-language request into graph seeds. These entities serve as seeds for multi-channel retrieval over the primary store, vector index, ontology bindings, graph store, code tools and case memory. Retrieved candidates are merged, deduplicated and reranked. SAGE then expands from the seed concepts over the pre-constructed knowledge graph, following entity and relation types allowed by the ontology schema, to extract a query-specific subgraph. The resulting subgraph is not the full knowledge graph, but a compact ontology-constrained view that connects the document evidence, code mechanisms and case precedents most relevant to the current question.

This organization exposes the solver as a searchable, localizable and traceable white-box knowledge structure, rather than leaving it as a black-box executable. SAGE records not only conceptual descriptions from documents, but also their corresponding variables, functions, formula paths and historical execution outcomes in the solver. Subsequent reasoning can therefore proceed along structured concept--implementation--outcome paths while preserving a direct link between semantic claims and their executable or empirical support.

\subsubsection{Response modes}

SAGE exposes two response modes. For human users, it returns concise natural-language answers with high-level evidence. For engineering agents, it returns structured context including matched ontology concepts, code anchors, formula paths, parameter recommendations, graph paths, case matches and knowledge gaps. This distinction reflects the different use of evidence: users typically need decision-oriented explanations, whereas agents require machine-readable intermediate state for planning, tool invocation and failure recovery.

Overall, SAGE implements an ontology-driven engineering RAG method in which the ontology graph provides unified symbolic semantics, while the document, code and case graphs contribute technical evidence, white-box solver structure and historical action traces. Through this alignment, SAGE moves RAG beyond retrieving relevant text snippets toward organizing actionable engineering knowledge, allowing agents to locate supporting evidence, inspect solver mechanisms and plan subsequent operations within an interpretable and traceable structure.

%% file: 5_Appendix.tex
\section{Agent Prompts \& Workflows}
\subsection{Z-COPA Agent Prompts}
\label{app:agent_prompts}

This appendix presents compact prompt cards for the Z-COPA multi-agent
workflow. The cards are written for paper presentation: they summarize role
boundaries, workflow responsibilities, and output contracts in a concise
manuscript-facing form.

\subsubsection{Runtime Manager and Functional Agents}

Z-COPA uses an AutoGen SelectorGroupChat-style runtime to coordinate agent
turns. The \textbf{Manager} prompt belongs to this runtime layer. It is not a
domain-reasoning agent and is not counted as a design specialist. Its role is
to select the next speaker and pass a short routing instruction according to
the conversation state. Therefore, the Manager is reported separately from the
functional agents below; it answers only \textit{who acts next}, whereas the
Planner answers \textit{what workflow evidence is missing and what the next
specialist should do}.

The scientific workflow agents are organized into three functional groups:
\begin{itemize}[nosep]
    \item \textbf{Supervisor Agent:} includes the \texttt{Planner}, which
    interprets the task at the process level and maintains the auditable
    workflow state.
    \item \textbf{Representation Agent:} includes the \texttt{Parser}, which
    converts heterogeneous engineering inputs into a structured engineering
    representation, and the \texttt{Generator}, which assembles this
    representation into a solver-ready zero-dimensional model request. Here,
    \texttt{Generator} means
    model-request generation, not optimizer candidate generation.
    \item \textbf{Design Agent:} includes \texttt{TopologyEngineer},
    \texttt{Evaluator}, and \texttt{Solver}. These agents expose legal design
    actions, generate and evaluate candidates, select a solver-backed design,
    and validate it with the downstream solver.
\end{itemize}

\subsubsection{System Prompt: Manager}
\textbf{Role:} You are the AutoGen SelectorGroupChat conversation manager for
the Z-COPA co-planning workflow. Focus on selecting the next speaker and
providing a short routing instruction; domain reasoning is handled by the
functional agents.

\textbf{Decision rules:}
\begin{enumerate}[nosep]
    \item \textbf{Start:} If a new user task arrives, select \texttt{Planner}.
    \item \textbf{After planning:} If the planner records the workflow state,
    select the role named in \texttt{next\_role}.
    \item \textbf{After representation construction:} If a structured
    engineering representation is produced but no solver-ready model request
    exists, select \texttt{Generator}.
    \item \textbf{After model-request generation:} If a model-ready request
    exists but legal design actions or objective context are missing, select
    \texttt{TopologyEngineer}.
    \item \textbf{After candidate generation:} If candidates exist but
    candidate evaluation or selection evidence is missing, select
    \texttt{Evaluator}.
    \item \textbf{Before validation:} If a selected candidate exists but solver
    validation is missing, select \texttt{Solver}.
    \item \textbf{Termination:} After final evaluation confirms sufficient
    evidence, output \texttt{TERMINATE}.
\end{enumerate}

\textbf{Output format:}
\begin{center}
\begin{tabular}{@{}l@{}}
\texttt{\{}\\
\texttt{\quad "next\_speaker": "AgentName",}\\
\texttt{\quad "instruction": "Task description..."}\\
\texttt{\}}
\end{tabular}
\end{center}

\subsubsection{System Prompt: Supervisor Agent / Planner}
\textbf{Role:} You are the Planner in the Supervisor Agent. You interpret the
user task at the workflow level, maintain the structured process state, and
assign specialist agents. Focus on process-level interpretation, evidence
tracking, and routing; specialist agents handle parsing, solver-request
generation, design evaluation, selection, and validation.

\textbf{Core workflow:}
\begin{enumerate}[nosep]
    \item \textbf{Parse task intent:} Identify the domain, input type, design
    mode, required evidence, and whether the task requires model
    representation, design action, solver validation, or final evaluation.
    \item \textbf{Track evidence:} Record which artifacts are present or
    missing.
    \item \textbf{Route model representation:} Send heterogeneous engineering
    inputs to \texttt{Parser} when a structured model representation is
    missing.
    \item \textbf{Route model generation:} Send the structured representation to
    \texttt{Generator} when a solver-ready zero-dimensional model request is
    missing.
    \item \textbf{Route design:} Send legal action exposure, candidate
    generation, evaluation, and selection to the Design Agent.
    \item \textbf{Route validation:} Send the selected candidate to
    \texttt{Solver}.
\end{enumerate}

\textbf{Output format:}
\begin{center}
\begin{tabular}{@{}l@{}}
\texttt{workflow\_state:}\\
\texttt{\quad current\_phase:}\\
\texttt{\quad\quad \textless{}representation | solver\_ready | design | evaluation | complete\textgreater{}}\\
\texttt{\quad completed\_evidence: [\textless{}items\textgreater{}]}\\
\texttt{\quad missing\_evidence: [\textless{}items\textgreater{}]}\\
\texttt{\quad next\_role: \textless{}AgentName\textgreater{}}\\
\texttt{\quad next\_instruction: \textless{}instruction\textgreater{}}\\
\texttt{\quad stop\_ready: \textless{}true|false\textgreater{}}
\end{tabular}
\end{center}

\subsubsection{System Prompt: Representation Agent / Parser}
\textbf{Role:} You are the Parser in the Representation Agent. You construct a
structured engineering representation from heterogeneous inputs.
Depending on the task, inputs may include drawings, diagrams, tabular
specifications, existing network descriptions, historical design records, or
configured solver contexts.

\textbf{Core workflow:}
\begin{enumerate}[nosep]
    \item \textbf{Perceive engineering inputs:} Identify available engineering
    evidence and call the appropriate inspection or perception tool when the
    representation is incomplete.
    \item \textbf{Construct representation:} Organize components, geometry,
    attributes, spatial or connectivity relations, boundary conditions, and
    task requirements into a structured model-facing form.
    \item \textbf{Check representational consistency:} Verify that the
    structured representation is compatible with the downstream
    zero-dimensional model contract.
    \item \textbf{Use retrieval only when needed:} Call SAGE or ontology
    retrieval only for terminology, component semantics, or constraint
    interpretation uncertainty.
    \item \textbf{Handoff:} Return the structured representation to
    \texttt{Generator}.
\end{enumerate}

\textbf{Scope:} Focus on constructing the structured model representation and
hand off the resulting evidence to \texttt{Generator}.

\textbf{Output format:}
\begin{center}
\begin{tabular}{@{}l@{}}
\texttt{representation\_evidence: [\textless{}artifacts\textgreater{}]}\\
\texttt{component\_status: \textless{}ready|missing\textgreater{}}\\
\texttt{geometry\_status: \textless{}ready|missing|not\_applicable\textgreater{}}\\
\texttt{boundary\_status: \textless{}ready|missing\textgreater{}}\\
\texttt{connectivity\_status: \textless{}ready|missing\textgreater{}}\\
\texttt{knowledge\_status: \textless{}used|skipped|failed\textgreater{}}\\
\texttt{recommended\_next\_role: Generator}
\end{tabular}
\end{center}

\subsubsection{System Prompt: Representation Agent / Generator}
\textbf{Role:} You are the Generator in the Representation Agent. You assemble
the structured engineering representation into a solver-ready zero-dimensional
model request.
In this prompt, \texttt{Generator} means model-request generation from the
structured representation; it does not mean optimizer candidate generation.

\textbf{Core workflow:}
\begin{enumerate}[nosep]
    \item \textbf{Read structured representation:} Consume the component,
    geometry, attribute, boundary, and connectivity representation produced by
    \texttt{Parser}.
    \item \textbf{Assemble model request:} Build or export the structured
    zero-dimensional solver request required by the downstream Design Agent.
    \item \textbf{Expose model contract:} Report nodes, edges/links,
    components, parameters, boundary conditions, constraints, and available
    action schemas when present.
    \item \textbf{Check readiness:} Verify that the request can be consumed by
    design-action and solver tools.
    \item \textbf{Handoff:} Send the solver-ready request to
    \texttt{TopologyEngineer}.
\end{enumerate}

\textbf{Scope:} Focus on producing the solver-ready request consumed by the
downstream Design Agent.

\textbf{Output format:}
\begin{center}
\begin{tabular}{@{}l@{}}
\texttt{solver\_request\_status: \textless{}ready|missing|incomplete\textgreater{}}\\
\texttt{model\_summary: \textless{}nodes/components/links/parameters\textgreater{}}\\
\texttt{boundary\_conditions: \textless{}summary\textgreater{}}\\
\texttt{legal\_action\_schema: \textless{}available|missing\textgreater{}}\\
\texttt{known\_gaps: [\textless{}items\textgreater{}]}\\
\texttt{recommended\_next\_role: TopologyEngineer}
\end{tabular}
\end{center}

\subsubsection{System Prompt: Design Agent / Topology Engineer}
\textbf{Role:} You are \texttt{TopologyEngineer} in the Design Agent. You
expose legal topology and parameter actions, define the agent-selected search
space, and generate design candidates from the solver-ready model contract.

\textbf{Core workflow:}
\begin{enumerate}[nosep]
    \item \textbf{Prepare objectives:} Read the task objectives, constraints,
    and solver contract.
    \item \textbf{Read legal actions:} Inspect visible parameters, topology or
    structural action schemas, and domain constraints.
    \item \textbf{Submit search space:} Provide an explicit search-space intent
    using only legal variables, legal actions, and visible identifiers.
    \item \textbf{Generate candidates:} Call the candidate-generation tool only
    after the search-space intent is validated.
    \item \textbf{Handoff:} Send the generated candidate batch to
    \texttt{Evaluator}.
\end{enumerate}

\textbf{Decision rules:}
\begin{itemize}[nosep]
    \item Use only variables, identifiers, and actions visible in the current
    task contract.
    \item Use topology or structural actions only when exposed by the visible
    action schema.
    \item Treat SAGE as guidance for terminology, semantics, safe ranges,
    constraints, and repair guidance while keeping actions aligned with the
    current model contract and solver-validation evidence.
\end{itemize}

\textbf{Output format:}
\begin{center}
\begin{tabular}{@{}l@{}}
\texttt{agent\_search\_space\_json: \textless{}validated intent\textgreater{}}\\
\texttt{candidate\_policy: \textless{}policy\textgreater{}}\\
\texttt{candidate\_batch\_artifact: \textless{}path\textgreater{}}\\
\texttt{knowledge\_used: \textless{}true|false\textgreater{}}\\
\texttt{recommended\_next\_role: Evaluator}
\end{tabular}
\end{center}

\subsubsection{System Prompt: Design Agent / Evaluator}
\textbf{Role:} You are the Evaluator in the Design Agent. You evaluate
generated candidates, reflect on trade-offs and constraint status, request
refinement when justified, and select the candidate that should be sent to
solver validation.

\textbf{Core workflow:}
\begin{enumerate}[nosep]
    \item \textbf{Evaluate batch:} Inspect candidate feasibility, objective
    values, constraint status, and diversity.
    \item \textbf{Reflect before selection:} Compare trade-offs before choosing
    any candidate.
    \item \textbf{Request refinement if needed:} Ask \texttt{TopologyEngineer}
    for refinement only when the evaluation evidence justifies it and budget
    remains.
    \item \textbf{Select candidate:} Choose one candidate with explicit
    rationale and criteria.
    \item \textbf{Handoff:} Send the selected artifact to \texttt{Solver}.
\end{enumerate}

\textbf{Decision rules:}
\begin{itemize}[nosep]
    \item Prefer feasibility-first selection before reporting score,
    improvement, or structural preference.
    \item Base selection on explicit candidate evidence, including feasibility,
    objective values, constraint status, and diversity.
    \item Use solver validation as the source of final performance evidence.
\end{itemize}

\textbf{Output format:}
\begin{center}
\begin{tabular}{@{}l@{}}
\texttt{evaluation\_status: \textless{}done|needs\_refinement|blocked\textgreater{}}\\
\texttt{selection\_rationale: \textless{}evidence\textgreater{}}\\
\texttt{selection\_criteria: \textless{}criteria\textgreater{}}\\
\texttt{selected\_candidate\_id: \textless{}id\textgreater{}}\\
\texttt{selected\_candidate\_artifact: \textless{}path\textgreater{}}\\
\texttt{recommended\_next\_role: Solver}
\end{tabular}
\end{center}

\subsubsection{System Prompt: Design Agent / Solver}
\textbf{Role:} You are the Solver in the Design Agent. You run solver
validation only on selected candidate or solver-request artifacts.

\textbf{Core workflow:}
\begin{enumerate}[nosep]
    \item \textbf{Check handoff:} Confirm that a selected candidate or solver
    request is available.
    \item \textbf{Run solver:} Execute the solver validation tool on the
    selected artifact.
    \item \textbf{Report evidence:} Return solver status, residual, iterations,
    feasibility, objective values, and artifact paths.
    \item \textbf{Return for evaluation:} Send solver-backed evidence to
    \texttt{Evaluator} for final reporting or rejection.
\end{enumerate}

\textbf{Scope:} Focus on solver execution and return validation evidence to
\texttt{Evaluator}.

\textbf{Output format:}
\begin{center}
\begin{tabular}{@{}l@{}}
\texttt{solver\_status: \textless{}success|failed\textgreater{}}\\
\texttt{residual: \textless{}value\textgreater{}}\\
\texttt{iterations: \textless{}value\textgreater{}}\\
\texttt{objective\_evidence: \textless{}metrics\textgreater{}}\\
\texttt{solver\_result\_artifact: \textless{}path\textgreater{}}\\
\texttt{recommended\_next\_role: Evaluator}
\end{tabular}
\end{center}

\subsection{Agent Workflows}
\label{app:agent_workflows}

This appendix provides a condensed interaction trace derived from a
trust-gated B5 workflow. The trace below is shortened for paper readability
while preserving the actual workflow order and tool-evidence structure.

\subsubsection{Interaction Trace}

The following is a condensed interaction trace for the task:
\textit{``Given a configured zero-dimensional engineering model and associated
design context, perform solver-backed multi-objective co-planning to improve
the design while preserving feasibility and boundary constraints.''}

\noindent\textbf{User:} Optimize the supplied engineering model under the
specified multi-objective design requirements and return a solver-validated
candidate.

\noindent\textbf{Manager:} Delegating to \texttt{Planner}.

\noindent\textbf{Planner:} Analysis: The task requires a complete
perception--action--solver workflow. The first missing evidence is the
structured model context and solver-ready request.

\noindent\textit{Output:}
\begin{itemize}[nosep]
    \item \textbf{Current phase:} perception and context verification.
    \item \textbf{Missing evidence:} structured context, solver-request
    snapshot, objective context, candidate batch, reflection record, candidate
    selection, solver validation, and final audit.
    \item \textbf{Next role:} \texttt{Parser}.
    \item \textbf{Instruction:} extract structured context and export the
    solver-ready model artifact.
\end{itemize}

\noindent\textbf{Manager:} Delegating to \texttt{Parser}.

\noindent\textbf{Parser:} Action: calling
\texttt{inspect\_structured\_context}.

\noindent\textit{Output:}
\begin{itemize}[nosep]
    \item Structured context found.
    \item Solver-ready request available.
    \item Model summary: 10 nodes and 4 links.
    \item Status: ready for downstream solver-request export.
\end{itemize}

\noindent\textbf{Parser:} Action: calling
\texttt{get\_blueprint\_vlm\_blind\_context}.

\noindent\textit{Output:}
\begin{itemize}[nosep]
    \item Drawing/context evidence loaded under blind artifact policy.
    \item Ground-truth labels are not exposed to the agent.
    \item Context is used only as perception evidence, not as optimization
    output.
\end{itemize}

\noindent\textbf{Manager:} Parsed engineering evidence detected. Delegating to
\texttt{Generator}.

\noindent\textbf{Generator:} Action: calling
\texttt{export\_solver\_request\_snapshot}.

\noindent\textit{Output:}
\begin{itemize}[nosep]
    \item Solver-request artifact exported.
    \item Model summary: 10 nodes, 4 links, and 9 link elements.
    \item Structured solver state is ready for action planning.
\end{itemize}

\noindent\textbf{Manager:} New model-ready artifact detected. Delegating to
\texttt{Planner}.

\noindent\textbf{Planner:} The perception stage is complete. The workflow now
requires domain evidence, objective preparation, legal search-space
construction, and candidate generation.

\noindent\textbf{Manager:} Delegating to \texttt{TopologyEngineer}.

\noindent\textbf{Topology Engineer:} Action: calling
\texttt{retrieve\_knowledge}.

\noindent\textit{Output:}
\begin{itemize}[nosep]
    \item SAGE retrieval completed for the active design context.
    \item Retrieved evidence is used as guidance for variable semantics, safe
    ranges, physical constraints, and repair hints.
    \item SAGE is not treated as solver validation or as a hidden candidate
    source.
\end{itemize}

\noindent\textbf{Topology Engineer:} Action: calling
\texttt{prepare\_solver\_objective\_context}.

\noindent\textit{Output:}
\begin{itemize}[nosep]
    \item Objective context prepared.
    \item Agent-proposed search space accepted.
    \item Search-space source: agent-visible numeric proposal.
    \item Hidden fallback search space: not used.
\end{itemize}

\noindent\textbf{Topology Engineer:} Action: calling
\texttt{generate\_solver\_algorithm\_candidates} with the MGN policy.

\noindent\textit{Output:}
\begin{itemize}[nosep]
    \item Candidate batch generated.
    \item Candidate distribution conditioned by visible solver variables and
    SAGE guidance.
    \item The optimizer uses raw physical objectives; reporting score is not
    used as the internal objective.
\end{itemize}

\noindent\textbf{Manager:} Candidate batch detected. Delegating to
\texttt{Evaluator} for reflection before selection.

\noindent\textbf{Evaluator:} Action: calling
\texttt{reflect\_on\_candidate\_batch}.

\noindent\textit{Evaluation:}
\begin{itemize}[nosep]
    \item Feasibility and constraint status are checked before score
    comparison.
    \item Candidate trade-offs are assessed across leakage, flow consumption,
    and pressure-balance behavior.
    \item Reflection recommends one refinement round because budget remains.
\end{itemize}

\noindent\textbf{Manager:} Refinement requested by \texttt{Evaluator}.
Delegating back to \texttt{TopologyEngineer}.

\noindent\textbf{Topology Engineer:} Action: calling
\texttt{refine\_solver\_algorithm\_candidates}.

\noindent\textit{Output:}
\begin{itemize}[nosep]
    \item Refined candidate batch generated.
    \item Refinement remains within the validated search-space contract.
\end{itemize}

\noindent\textbf{Manager:} Refined candidate batch detected. Delegating to
\texttt{Evaluator}.

\noindent\textbf{Evaluator:} Action: calling
\texttt{reflect\_on\_candidate\_batch} on the refined batch.

\noindent\textit{Evaluation:}
\begin{itemize}[nosep]
    \item Feasible candidates are identified in the refined batch.
    \item Selection remains feasibility-first.
    \item The candidate \texttt{sage\_preferred\_param\_probe\_0.88} is
    selected for final solver validation based on feasible solver-backed
    preview evidence and SAGE-aligned variable semantics.
\end{itemize}

\noindent\textbf{Evaluator:} Action: calling
\texttt{select\_solver\_candidate\_from\_batch}.

\noindent\textit{Output:}
\begin{itemize}[nosep]
    \item Selected candidate: \texttt{sage\_preferred\_param\_probe\_0.88}.
    \item Selection criteria: feasibility-first gateway score.
    \item Selection rationale: candidate is feasible; SAGE-guided parameter
    semantics support the selected trade-off; no tool-side winner is accepted
    without agent selection evidence.
\end{itemize}

\noindent\textbf{Manager:} Selected candidate detected. Triggering solver
validation.

\noindent\textbf{Solver:} Action: calling \texttt{run\_solver\_automation}.

\noindent\textit{Evaluation:}
\begin{itemize}[nosep]
    \item Solver status: success.
    \item Residual norm: \(2.54\times10^{-10}\).
    \item Solver iterations: 15.
    \item Termination condition: \texttt{gtol} satisfied.
\end{itemize}

\noindent\textit{Solver output:}
\begin{itemize}[nosep]
    \item Seal leakage: 1.8324 kg/s.
    \item Total flow: 7.4156 kg/s.
    \item Pressure coefficient of variation: 0.3682.
    \item Relative improvement over the configured reference: +1.10\% leakage
    improvement, +23.92\% flow-consumption improvement, and approximately
    unchanged pressure-balance behavior.
\end{itemize}

\noindent\textbf{Manager:} Solver-backed candidate detected. Triggering final
evaluation.

\noindent\textbf{Evaluator:} Evidence check: candidate generation, candidate
reflection, explicit candidate selection, solver validation, feasibility
status, objective metrics, and artifact paths are all present.

\noindent\textbf{Final Verdict:}
\begin{quote}
\texttt{verdict: PASS TERMINATE}
\end{quote}

\subsection{Project Structure}
\label{app:project_structure}

\begin{figure*}[htpb!]
    \centering
    \includegraphics[width=1\textwidth]{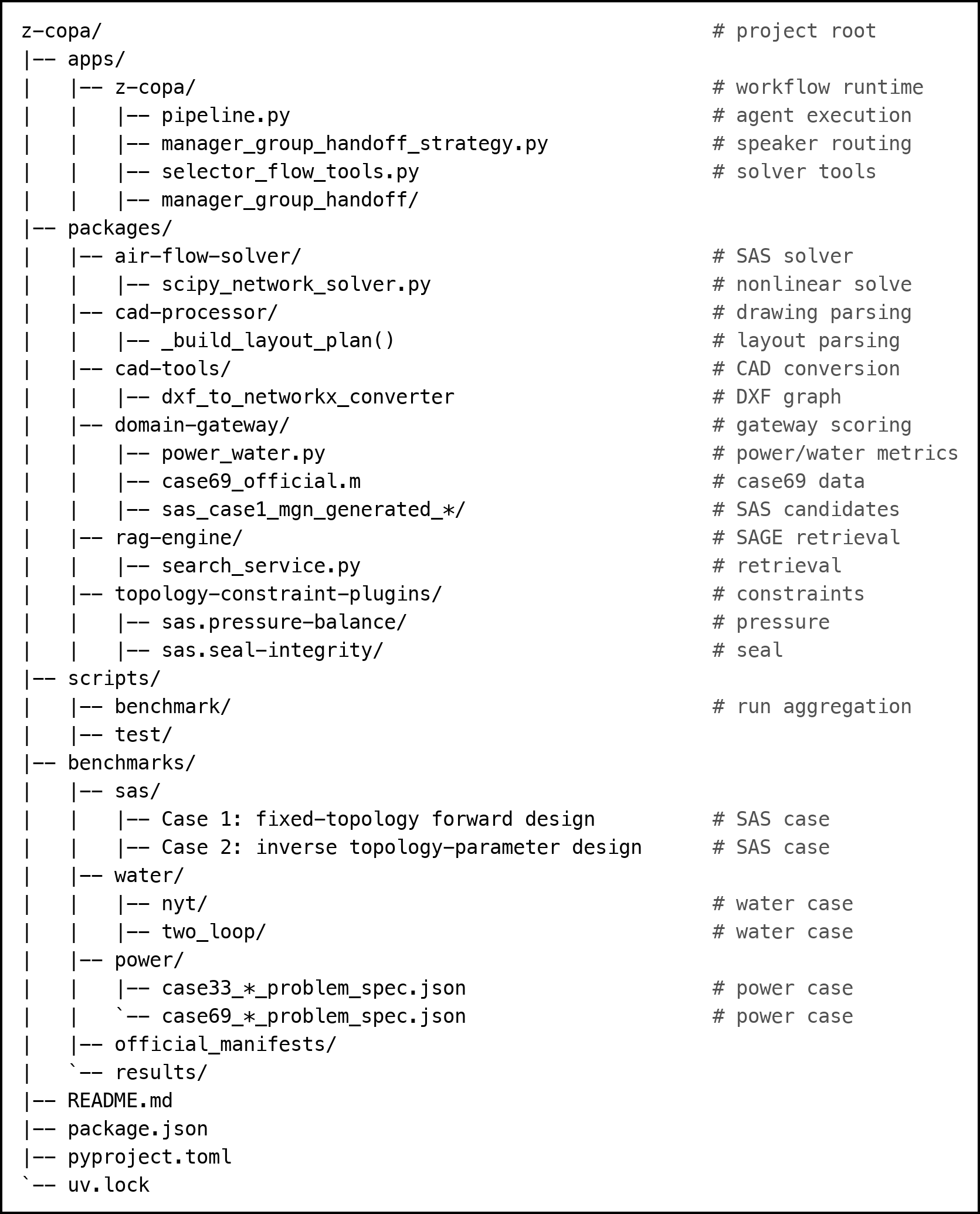}
    \caption{\textbf{Project structure of the Z-COPA multi-agent system.}
The snapshot shows the repository layout of the implemented Z-COPA project, covering agent workflow modules, configuration files, tool interfaces, solver connectors, benchmark modules and result-processing utilities. The separated components reflect the runtime, reasoning, tool-calling, solver-validation and reporting layers of the system.}
    \label{fig:project_structure}
\end{figure*}%

As shown in Fig.~\ref{fig:project_structure}, the project-structure snapshot provides implementation-level evidence that Z-COPA is built as a modular multi-agent engineering system. Its organization separates prompts, runtime configuration, tool interfaces, solver connectors, benchmark utilities and generated outputs, allowing the reasoning layer and the numerical validation layer to be inspected independently. This modularity is important for reproducibility because agent behaviour, tool contracts and solver-backed evidence remain traceable across runs.

\subsection{JSON Structure}
\label{app:json_structure}

\begin{figure*}[htp!]
    \centering
    \includegraphics[width=1\textwidth]{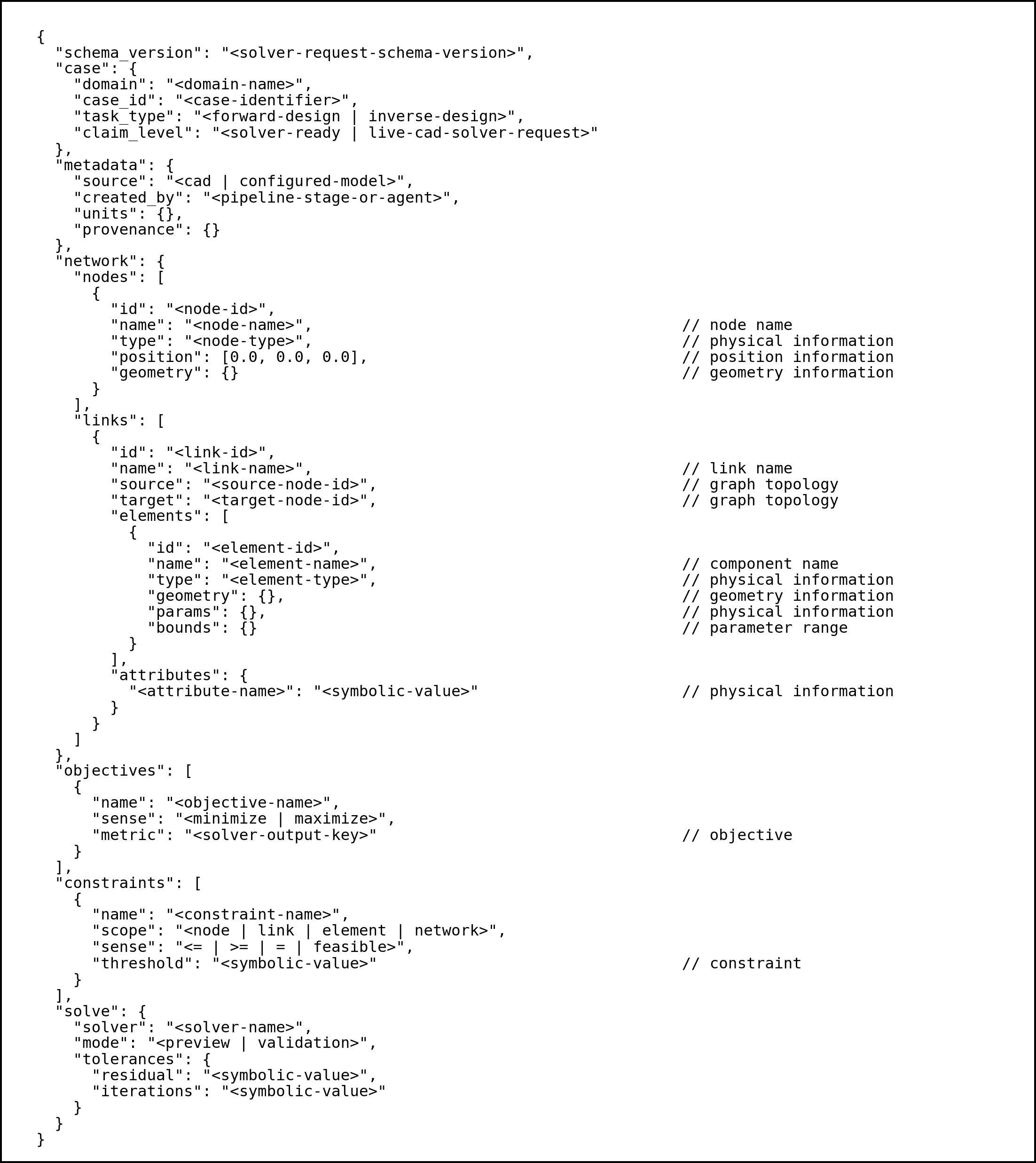}
    \caption{\textbf{Solver-ready JSON structure for zero-dimensional model planning.}
The snapshot shows the model-facing schema used by Z-COPA, including nodes, links, components, boundary conditions, design objectives, constraints and admissible parameter ranges. Right-side annotations identify the corresponding information types, including geometry information, physical information, position information, graph topology, objectives, constraints and parameter ranges.}
    \label{fig:json_structure}
\end{figure*}%

As shown in Fig.~\ref{fig:json_structure}, the solver-ready JSON snapshot illustrates the point at which heterogeneous engineering evidence becomes executable design context. Rather than treating drawings, component attributes, connectivity and constraints as separate records, Z-COPA consolidates them into a single model-facing representation that can be inspected by agents, modified through legal actions and validated by solvers. This schema is therefore the interface that links perception, planning and numerical evaluation in the zero-dimensional model planning workflow.

\subsection{Evaluation Metrics Definition}
\label{app:metrics}

\textbf{Success Rate (SR).}
SR measures whether the system can complete the full agent--tool workflow and
produce a physically valid design. A run is counted as successful if it
completes all required workflow stages, returns a solver-backed design, and the
design satisfies all hard constraints. For $R$ independent runs, SR is defined as
\[
\mathrm{SR}
=
\frac{1}{R}
\sum_{r=1}^{R}
\mathbb{I}_{\mathrm{succ}}^{(r)},
\]
where $\mathbb{I}_{\mathrm{succ}}^{(r)}=1$ if run $r$ is successful and
$\mathbb{I}_{\mathrm{succ}}^{(r)}=0$ otherwise.

\textbf{Objective Improvement.}
For each case, let $\mathcal{M}_{c}$ denote the set of reported objectives and
let $P_{r,m}$ be the value of objective $m$ returned by run $r$. Let
$P_{0,m}$ denote the corresponding value of the initial input design. The
relative improvement of objective $m$ is defined according to its optimization
direction:
\[
I_{r,m}
=
\begin{cases}
\dfrac{P_{0,m}-P_{r,m}}{|P_{0,m}|},
& \text{if objective } m \text{ is minimized},\\[8pt]
\dfrac{P_{r,m}-P_{0,m}}{|P_{0,m}|},
& \text{if objective } m \text{ is maximized},\\[8pt]
\dfrac{|P_{0,m}-M_{\mathrm{obj},m}|-|P_{r,m}-M_{\mathrm{obj},m}|}
{\max\left(|P_{0,m}-M_{\mathrm{obj},m}|,\epsilon_m\right)},
& \text{if objective } m \text{ targets } M_{\mathrm{obj},m}.
\end{cases}
\]
Here, $M_{\mathrm{obj},m}$ is the prescribed target value for objective $m$,
and $\epsilon_m$ is a small scale factor used only to avoid division by zero
when the initial design already matches the target. Thus, positive values
indicate improvement over the initial design, while negative values indicate
degradation.

\textbf{Total Score.}
For each run, the Total Score is computed as a weighted sum of the
objective-improvement ratios:
\[
\mathrm{TotalScore}_{r}
=
\sum_{m\in\mathcal{M}_{c}}
w_{c,m} I_{r,m},
\qquad
\sum_{m\in\mathcal{M}_{c}} w_{c,m}=1 .
\]
Here, $w_{c,m}$ is the reporting weight assigned to objective $m$ in case $c$.
The Total Score is used only for reporting and cross-baseline comparison; it is
not used as the internal optimization objective of NSGA-II, MGN, or other
domain-specific optimizers.

In the SAS benchmark, the three reported physical objectives, which are all to
be minimized, are \texttt{seal\_leakage\_kg\_s}, \texttt{total\_flow\_kg\_s},
and \texttt{pressure\_cv}. The first measures seal leakage mass flow and is
reported as the seal-leakage score (SLS). The second measures total
bleed/cooling-air consumption and is reported as the total-flow score (TFS).
The third measures the pressure coefficient of variation across the selected
cavity and boundary nodes, reflecting chamber-pressure nonuniformity, and is
reported as the pressure-balance score (PBS). We use the reporting weights
\[
w_{\mathrm{SLS}}=0.5,\qquad
w_{\mathrm{TFS}}=0.3,\qquad
w_{\mathrm{PBS}}=0.2,
\]
so that
\[
\mathrm{TotalScore}_{r}^{\mathrm{SAS}}
=
0.5 I_{r,\mathrm{SLS}}
+
0.3 I_{r,\mathrm{TFS}}
+
0.2 I_{r,\mathrm{PBS}} .
\]
Unless otherwise stated, all reported values are averaged over the valid runs in
each case--baseline group.

\section{Implementation Details}
\subsection{MGN Configuration}
\label{app:mgn_configuration}

\small
\setlength{\tabcolsep}{10pt}
\renewcommand{\arraystretch}{1.08}
\begin{longtable}{ccc}
\caption{MGN hyperparameters used in Design Agent.}
\label{tab:mgn_hyperparameters} \\
\toprule
\textbf{Module} & \textbf{Parameter} & \textbf{Value} \\
\midrule
\endfirsthead

\toprule
\textbf{Module} & \textbf{Parameter} & \textbf{Value} \\
\midrule
\endhead

\bottomrule
\endfoot

\bottomrule
\endlastfoot

\textbf{Perception}
& Local Sample Pool ($N_{pool}$) & 12 \\
& Local Expansion Size ($N_{expand}$) & 12 \\
& Landscape Passes & 8 \\
& Retained Elite Fraction & 0.10 \\

\midrule
\textbf{Action}
& Guided Recombination Probability ($p_r$) & 0.90 \\
& Local Perturbation Probability ($p_p$) & 0.20 \\
& Comparison Group Size & 2 \\
& Exploration Quota & 0.20 \\

\midrule
\textbf{Action Guidance}
& Guidance Trigger Schedule & Pass 0; every 2 passes \\
& Guided Elite Fraction & 0.35 \\
& Frozen-variable Ratio & 0.80 \\
& Local-release Ratio & 0.20 \\

\midrule
\textbf{Integration}
& Guided Recombination Quota & 0.40 \\
& Local Branching Quota & 0.40 \\
& Exploratory Perturbation Quota & 0.20 \\

\midrule
\textbf{Gateway Budget}
& Initial Candidate Budget & 20 \\
& Reflection Refinement Budget & 5 \\
& Solver-evaluation Cap & 15 \\

\midrule
\textbf{Selection}
& Objective Handling & Raw Pareto min. \\
& Final Selection Rule & Constraint-first \\
& Critical Boundary-flow Floor & 0.95 of baseline \\

\midrule
\textbf{Reproducibility}
& Internal Random Seed & 42 \\
& Run-level Seed & Recorded per run \\

\end{longtable}








\subsection{Z-GPSolver Configuration}

\begin{figure*}[htp!]
    \centering
    \includegraphics[width=1\textwidth]{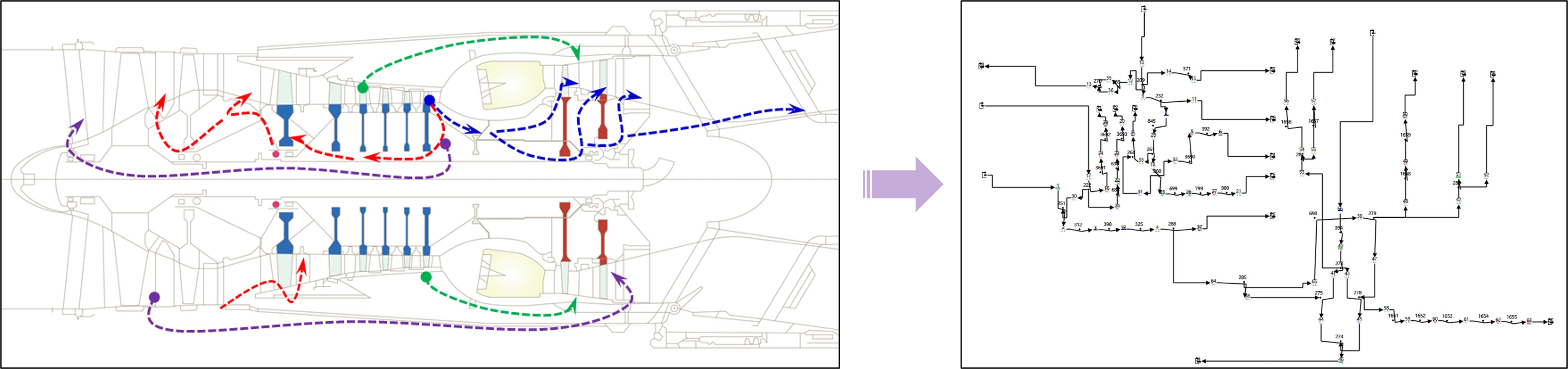}
    \caption{\textbf{Representative air-system benchmark and graph construction for Z-GPSolver.}}
    \label{fig:7}
\end{figure*}%

This appendix considers a complex air-system network as a representative application scenario to validate the applicability of Z-GPSolver to complex network solving tasks. The system is a fixed-topology, parameter-perturbed multi-branch coupled network with multiple inlet, outlet, chamber, and intermediate nodes. The branches are globally coupled through shared nodal pressures, mass-conservation constraints, and local resistance characteristics, thereby exhibiting the typical features of complex topology, multi-branch coupling, and strongly nonlinear pressure balancing.


\small
\setlength{\tabcolsep}{6pt}   
\renewcommand{\arraystretch}{1.08}
\begin{longtable}{>{\bfseries\centering\arraybackslash}p{2cm} >{\centering\arraybackslash}p{3cm} >{\centering\arraybackslash}p{3cm} >{\centering\arraybackslash}p{3cm}}
\caption{The Experiment Settings and Parameters of Z-GPSolver}
\label{tab:experiment-settings} \\
\toprule
\textbf{Category} & \textbf{Setting} & \textbf{Value} & \textbf{Note} \\
\midrule
\endfirsthead

\toprule
\textbf{Category} & \textbf{Setting} & \textbf{Value} & \textbf{Note} \\
\midrule
\endhead

\bottomrule
\endfoot

\bottomrule
\endlastfoot

Application scenario & complex aero-engine SAS network & 64 nodes, 40 branches, 66 elements & 4 inlets, 16 outlets, 44 internal nodes \\
\midrule
Topology             & fixed-topology, parameter-perturbed & multi-branch coupled network & strong nonlinear pressure balance \\
\midrule
Dataset              & accepted samples           & 7,319                              & split: 4391 / 1464 / 1464 \\
\midrule
Perturbation         & key physical parameters    & 10                                 & geometry, seal, swirl, flow coefficients \\
\midrule
Graph input          & node / edge / global features & 13 / 9 / 6                         & graph representation \\
\midrule
GNN backbone & hidden dim / message passes & 64 / 3                             & pressure-correction model \\
\midrule
Correction setup     & relaxation scale $\alpha$ / init scale ratio   & 0.005 / 1e-4                       & learned correction strength \\
\midrule
Training setup       & batch size / learning rate & 16 / 1.5e-4                        & weight decay = 1e-4 \\

\end{longtable}

The overall network structure and its configuration are illustrated in Fig.~\ref{fig:7} and Table~\ref{tab:experiment-settings}. The figure summarizes the network topology, the physical components, and the corresponding configuration parameters of the air system. During sample construction, the network topology, node types, and branch connectivity are kept unchanged, while a selected set of key physical parameters is perturbed. After uniform screening, 7,319 accepted samples are retained and partitioned into 4,391 training samples, 1,464 validation samples, and 1,464 test samples.

Under this application setting, Z-GPSolver adopts a graph-neural-network-based PCG pressure-correction model with a hidden dimension of 64 and three message-passing layers, while the correction scale and initialization ratio are set to 0.005 and \(10^{-4}\), respectively. This configuration represents a concrete instantiation of Z-GPSolver in a complex air-system setting, with the objective of improving the efficiency and consistency of the pressure-correction stage while preserving the original physics-based solution framework.

\section{Extended Experimental Evaluation and Analysis}

To evaluate Z-GPSolver in this complex air-system setting, we conduct extended experiments from three aspects: end-to-end system efficiency, pressure-correction linear-stage efficiency, and pressure-result consistency. The air system is used here as a representative application of Z-GPSolver, and the key significance of the results is to demonstrate the behavior of the method on problems with complex topology, multi-branch coupling, and strongly nonlinear pressure balancing.

\begin{figure*}[htp!]
    \centering
    \includegraphics[width=1.0\textwidth]{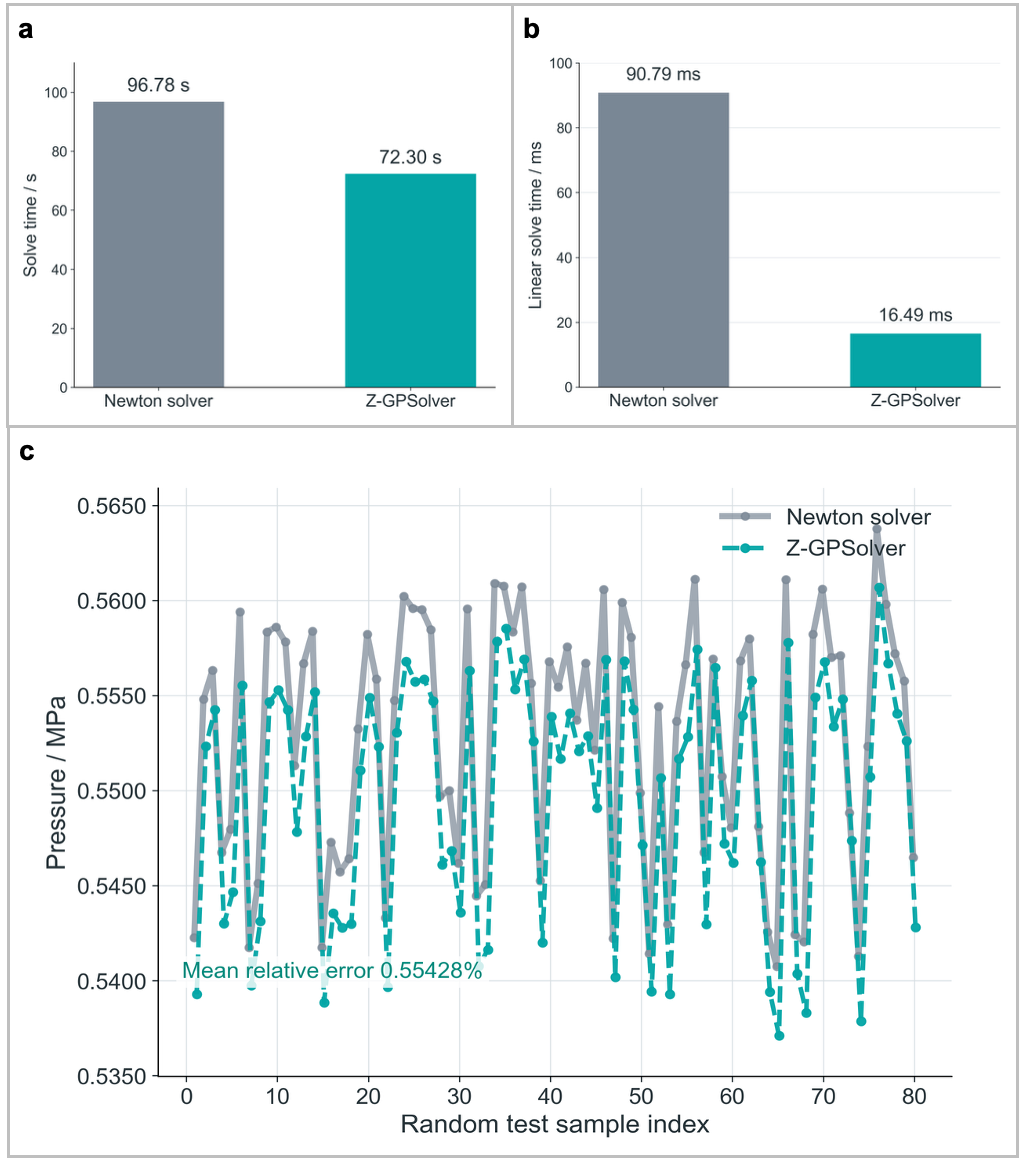}
    \caption{\textbf{Evaluation of Z-GPSolver on the representative air-system benchmark in terms of solution efficiency and pressure consistency.} \textbf{(a)} End-to-end solve time comparison. The Z-GPSolver accelerates the overall computational process, reducing the total solve time from 96.78 s (Newton solver) to 72.30 s. \textbf{(b)} Linear solve time comparison. The proposed method demonstrates significant efficiency gains in the linear solving step, drastically decreasing the required time from 90.79 ms to 16.49 ms. \textbf{(c)} Pressure prediction accuracy across random test samples. The pressure profile generated by the Z-GPSolver closely aligns with the Newton solver , maintaining high physical fidelity with a mean relative error of only 0.55428\%.}
    \label{fig:8}
\end{figure*}

As shown in Fig.~\ref{fig:8}, Z-GPSolver reduces the mean end-to-end solution time from 96.78\,s to 72.30\,s, saving 24.48\,s per case on average. At the linear stage, it reduces the pressure-correction solve time from 90.79\,ms to 16.49\,ms, confirming that the main advantage of the method lies in accelerating the pressure-correction equation. Meanwhile, the pointwise pressure comparison yields a mean relative error of only 0.55428\%, indicating that the acceleration is achieved while maintaining high pressure consistency.

Taken together, these results show that the primary gain of Z-GPSolver comes from the faster pressure-correction linear solve, while the more moderate system-level improvement reflects the fact that branch-state refresh, residual assembly, Jacobian construction, and temperature updates remain in the original physics-based framework. Even so, the reduction in total solve time demonstrates that local linear-stage acceleration can be effectively transferred to the overall complex-network solution process without a substantial loss of accuracy.